%% file: main.tex
\documentclass[letterpaper, 10 pt, conference]{ieeeconf}

\IEEEoverridecommandlockouts %
\usepackage[utf8]{inputenc}
\usepackage[english]{babel}
\usepackage{amsmath}
\usepackage{amssymb}
\usepackage{mathtools}
\usepackage{amsfonts}
\usepackage[hidelinks]{hyperref}
\usepackage[acronym,abbreviations]{glossaries-extra}
\usepackage{glossaries-prefix}
\usepackage[capitalise]{cleveref}
\usepackage{textcomp}
\usepackage{tikz}
\usepackage{float}
\usepackage{array}
\usepackage{import}
\usepackage{MnSymbol}
\usetikzlibrary{arrows,automata}
\usepackage{algorithm}
\usepackage{algpseudocode}
\usepackage{algorithmicx}
\usepackage{bbm}
\usepackage{xparse}
\usepackage{pgfplots}
\usepackage{pgfplotstable}
\usepackage{wrapfig}
\usepackage{graphicx}
\usepackage{subcaption}
\usepgfplotslibrary{fillbetween}
\usepackage{xcolor}
\usepackage{autonum}
\usepackage[font=small]{caption}
\usepackage{microtype}

\cslet{blx@noerroretextools}\empty
\usepackage[backend=biber,style=ieee,natbib=true,citestyle=numeric-comp]{biblatex}

\bibliography{bibliography/active_inference}
\bibliography{bibliography/general}
\bibliography{bibliography/mi}
\bibliography{bibliography/model_based_rl}
\bibliography{bibliography/active_inference_related_work}
\bibliography{bibliography/boed_related_work}
\bibliography{bibliography/appendix}

\include{def}

\pgfplotsset{rewplot/.style={
    align=center,
    legend style={nodes={scale=0.6, transform shape}},
    grid=both,
    width=1.1\textwidth,
    height=4.2cm,
    reverse legend,
    legend pos=north west,
    legend cell align={left},
    scaled x ticks=base 10:-3,
    y label style={at={(axis description cs:0.1,.5)}},
    xlabel = {\footnotesize Steps},
    ylabel = {~}
}}

\setabbreviationstyle[acronym]{long-short}

\newcommand{\anaacr}[3]{%
    \newacronym[prefixfirst={a\ },prefix={an\ }]{#1}{#2}{#3}%
}

\anaacr{fep}{FEP}{Free Energy Principle}
\newacronym{vfe}{VFE}{Variational Free Energy}
\newacronym{efe}{EFE}{Expected Free Energy}
\anaacr{fef}{FEF}{Free Energy of the Future}
\newacronym{gfe}{GFE}{Generalized Free Energy}
\anaacr{feef}{FEEF}{Free Energy of the Expected Future}
\newacronym{ai}{AI}{Active Inference}
\newacronym{rl}{RL}{Reinforcement Learning}
\newacronym{pomdp}{POMDP}{Partially Observable Markov Decision Process}
\anaacr{mdp}{MDP}{Markov Decision Process}
\anaacr{mpc}{MPC}{Model Predictive Control}
\newacronym{cem}{CEM}{Cross-Entropy Method}
\anaacr{mc}{MC}{Monte Carlo}
\anaacr{mi}{MI}{Mutual Information}
\newacronym{li}{LI}{Lautum Information}
\anaacr{nmc}{NMC}{Nested Monte Carlo}
\anaacr{sgd}{SGD}{Stochastic Gradient Descent}
\anaacr{svgd}{SVGD}{Stein Variational Gradient Descent}
\newacronym{rbf}{RBF}{Radial Basis Function}
\newacronym{boed}{BOED}{Bayesian Optimal Experimental Design}
\newacronym{elbo}{ELBO}{Evidence Lower Bound}
\anaacr{mse}{MSE}{Mean Squared Error}
\newacronym{sac}{SAC}{Soft Actor Critic}
\anaacr{mcts}{MCTS}{Monte Carlo Tree Search}
\anaacr{mtl}{MTL}{multi-task learning}

\newacronym{em}{EM}{Expectation Maximization}
\glsunset{em}
\newacronym{iid}{i.i.d.}{independent and identically distributed}
\glsunset{iid}
\newacronym{kl}{KL}{Kullback-Leibler}
\glsunset{kl}

\makeglossaries

\DeclareUnicodeCharacter{1EF3}{\`y}

\newcommand{\balltaskname}{Tilted Pushing}
\newcommand{\ballhtaskname}{Tilted Pushing Maze}
\newcommand{\ballrtaskname}{Tilted Pushing Real}

\makeatletter
\newcommand*\myfontsize{%
  \@setfontsize\myfontsize{7.6}{9.0}%
}
\makeatother

\def\fullversion{1}

\title{\LARGE \bf Active Exploration for Robotic Manipulation}

\author{
    Tim Schneider$^{1}$, Boris Belousov$^{1}$, Georgia Chalvatzaki$^{1}$, Diego Romeres$^{2}$, Devesh K. Jha$^{2}$ and Jan Peters$^{1}$
    \thanks{$^{1}$Tim Schneider, Boris Belousov, Georgia Chalvatzaki, and Jan Peters are with the Intelligent Autonomous Systems Lab, Technical University of Darmstadt, 64289 Darmstadt, Germany, {\tt\small \{tim.schneider1, name.surname\}@tu-darmstadt.de}}%
    \thanks{$^{2}$Diego Romeres, Devesh K. Jha are with the Mitsubishi Electric Research Laboratories (MERL), Cambridge, MA, USA, {\tt\small \{romeres,jha\}@merl.com}}%
    \thanks{This project has received funding from BMWSB ZukunftBau under grant Nr. 10.08.18.7-21.34 and the funded programs Aristotle by BMBF and iROSA by DFG. Calculations for this research were partially conducted on the Lichtenberg high performance computer of the TU Darmstadt.}
}

\glsdisablehyper

\begin{document}
    \maketitle
    \thispagestyle{empty}
    \pagestyle{empty}

    \begin{abstract}

        Robotic manipulation stands as a largely unsolved problem despite significant advances in robotics and machine learning in recent years.
        One of the key challenges in manipulation is the exploration of the dynamics of the environment when there is continuous contact between the objects being manipulated.
        This paper proposes a model-based active exploration approach that enables efficient learning in sparse-reward robotic manipulation tasks.
        The proposed method estimates an information gain objective using an ensemble of probabilistic models and deploys model predictive control (MPC) to plan actions online that maximize the expected reward while also performing directed exploration.
        We evaluate our proposed algorithm in simulation and on a real robot, trained from scratch with our method, on a challenging ball pushing task on tilted tables, where the target ball position is not known to the agent a-priori. 
        Our real-world robot experiment serves as a fundamental application of active exploration in model-based reinforcement learning of complex robotic manipulation tasks. 
        Project page \url{https://sites.google.com/view/aerm}.

    \end{abstract}

    \input{s0_introduction}

    \input{s1_related_work}

    \input{s2_method}

    \input{s3_experiments.tex}

    \input{s4_active_inference.tex}

    \input{s5_conclusion}

    \printbibliography[notkeyword=appendix]

    \ifnum1=\fullversion
        \input{s6_appendix.tex}
    \fi
\end{document}

%% file: def.tex
\DeclarePairedDelimiterX{\infdivx}[2]{[}{]}{%
    #1\delimsize\|\;#2%
}
\newcommand{\kl}{D_{\text{KL}}\infdivx}

\newcommand{\psym}{p}
\newcommand{\Psym}{P}
\newcommand{\qsym}{q}
\newcommand{\Qsym}{Q}
\newcommand{\pssym}{\psym^\ast}
\newcommand{\qssym}{\qsym^\ast}

\NewDocumentCommand\pr{mmo}{%
    \ensuremath{%
        #1\mathchoice{\!}{\!}{}{}\left( #2 %
        \IfNoValueTF{#3}{}{\,\middle\lvert\, #3} %
        \right)%
    }
}

\NewDocumentCommand\prs{ommo}{%
    \ensuremath{\pr{\IfNoValueTF{#1}{#2}{#2_{#1}}}{#3}[#4]}
}

\NewDocumentCommand\p{omo}{\prs[#1]{\psym}{#2}[#3]}
\NewDocumentCommand\ps{omo}{\prs[#1]{\pssym}{#2}[#3]}
\NewDocumentCommand\PP{omo}{\prs[#1]{\Psym}{#2}[#3]}
\NewDocumentCommand\q{omo}{\prs[#1]{\qsym}{#2}[#3]}
\NewDocumentCommand\Q{omo}{\prs[#1]{\Qsym}{#2}[#3]}
\NewDocumentCommand\qs{omo}{\prs[#1]{\qssym}{#2}[#3]}

\newcommand{\vs}{s}
\newcommand{\vh}{x}
\newcommand{\vo}{o}
\newcommand{\vr}{r}
\newcommand{\va}{a}
\newcommand{\vte}{\mathcal{T}}

\newcommand{\misym}{\text{MI}}
\newcommand{\lisym}{\text{LI}}
\NewDocumentCommand\mi{mmo}{\pr{\misym}{#1, #2}[#3]}
\NewDocumentCommand\li{mmo}{\pr{\lisym}{#1, #2}[#3]}

\newcommand{\R}{\mathbb{R}}
\newcommand{\E}{\mathbb{E}}
\newcommand{\Ex}[2]{\E_{#1}\mathchoice{\!}{\!}{}{}\left[ #2 \right]}

\DeclareMathOperator*{\argmax}{arg\,max}

\addto\extrasenglish{%
    \crefname{approx}{Approximation}{Approximations}%
  \Crefname{approx}{Approximation}{Approximations}%
  \creflabelformat{approx}{#2(#1)#3}
}

\addto\extrasenglish{%
  \crefname{constr}{Constraint}{Constraints}%
  \Crefname{constr}{Constraint}{Constraints}%
  \creflabelformat{constr}{#2(#1)#3}
}

\addto\extrasenglish{%
  \crefname{def}{Definition}{Definitions}%
  \Crefname{def}{Definition}{Definitions}%
  \creflabelformat{def}{#2(#1)#3}
}

\newcommand{\vfesym}{\mathcal{F}}

\newcommand{\efesymP}{G}

\pgfplotsset{
    every axis plot/.append style={line width=0.8pt},
    every axis plot post/.append style={
        every mark/.append style={mark=none}
    }
}

\definecolor{pltBlue}{HTML}{1f77b4}
\definecolor{pltOrange}{HTML}{ff7f0e}
\definecolor{pltGreen}{HTML}{2ca02c}
\definecolor{pltRed}{HTML}{d62728}
\definecolor{pltPurple}{HTML}{9467bd}
\definecolor{pltBrown}{HTML}{8c564b}
\definecolor{pltPink}{HTML}{e377c2}
\definecolor{pltGray}{HTML}{7f7f7f}
\definecolor{pltBeige}{HTML}{bcbd22}
\definecolor{pltLightBlue}{HTML}{17becf}

\NewDocumentCommand\plotWithStddev{mmmmmmmO{1}}{\plotstd{#1}{#2}{#3}{#4}{#5}{#6}[#7][#8]}

\NewDocumentCommand\plotstd{mmmmmmoO{1}O{1}}{%
    \pgfplotstableread[col sep=comma]{#1}\datatable
    \IfNoValueTF{#7}{}{%
        \addplot[smooth, opacity=0, name path=A, forget plot] table [x expr=#9 * \thisrow{#5}, y expr=#3 * (\thisrow{#6} - #8 * \thisrow{#7})] {\datatable};
        \addplot[smooth, opacity=0, name path=B, forget plot] table [x expr=#9 * \thisrow{#5}, y expr=#3 * (\thisrow{#6} + #8 * \thisrow{#7})] {\datatable};
        \addplot[#2, opacity=0, fill opacity=0.4]  fill between [of=A and B];
        \addlegendentry{$\pm$ std};
    }
    \addplot+[color=#2, smooth, solid] table [x expr=#9 * \thisrow{#5}, y expr=#3 * \thisrow{#6}] {\datatable};
    \addlegendentry{#4};
}

\NewDocumentCommand\plotstdcom{mmmmmmoO{1}O{1}}{%
    \pgfplotstableread[col sep=comma]{#1}\datatable
    \IfNoValueTF{#7}{}{%
        \addplot[smooth, opacity=0, name path=A, forget plot] table [x expr=#9 * \thisrow{#5}, y expr=#3 * (\thisrow{#6} - #8 * \thisrow{#7})] {\datatable};
        \addplot[smooth, opacity=0, name path=B, forget plot] table [x expr=#9 * \thisrow{#5}, y expr=#3 * (\thisrow{#6} + #8 * \thisrow{#7})] {\datatable};
        \addplot[#2, opacity=0, fill opacity=0.4, forget plot]  fill between [of=A and B];
    }
    \addplot+[color=#2, smooth, solid] table [x expr=#9 * \thisrow{#5}, y expr=#3 * \thisrow{#6}] {\datatable};
    \addlegendentry{#4};
}

\newcolumntype{C}{>{\centering\arraybackslash}m{0.11\linewidth}}
\newcolumntype{D}{>{\centering\arraybackslash}m{0.10\linewidth}}

%% file: s0_introduction.tex
\section{Introduction}

\global\csname @topnum\endcsname 0
\global\csname @botnum\endcsname 0

\begin{figure}
    \centering
    \begin{subfigure}[t]{0.49\linewidth}
        \centering
        \includegraphics[width=\linewidth]{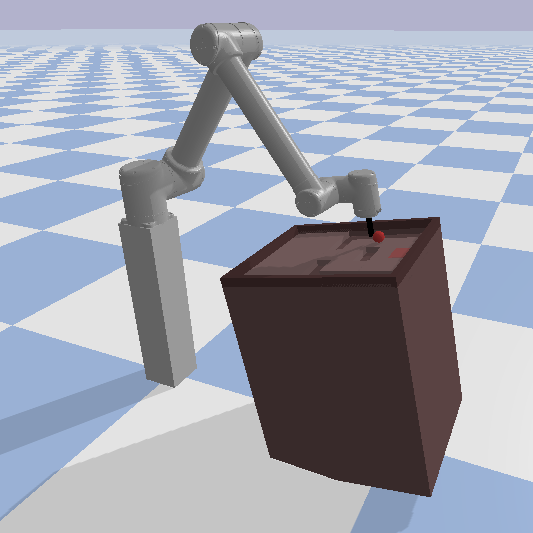}
    \end{subfigure}
    \begin{subfigure}[t]{0.49\linewidth}
        \centering
        \includegraphics[width=\linewidth]{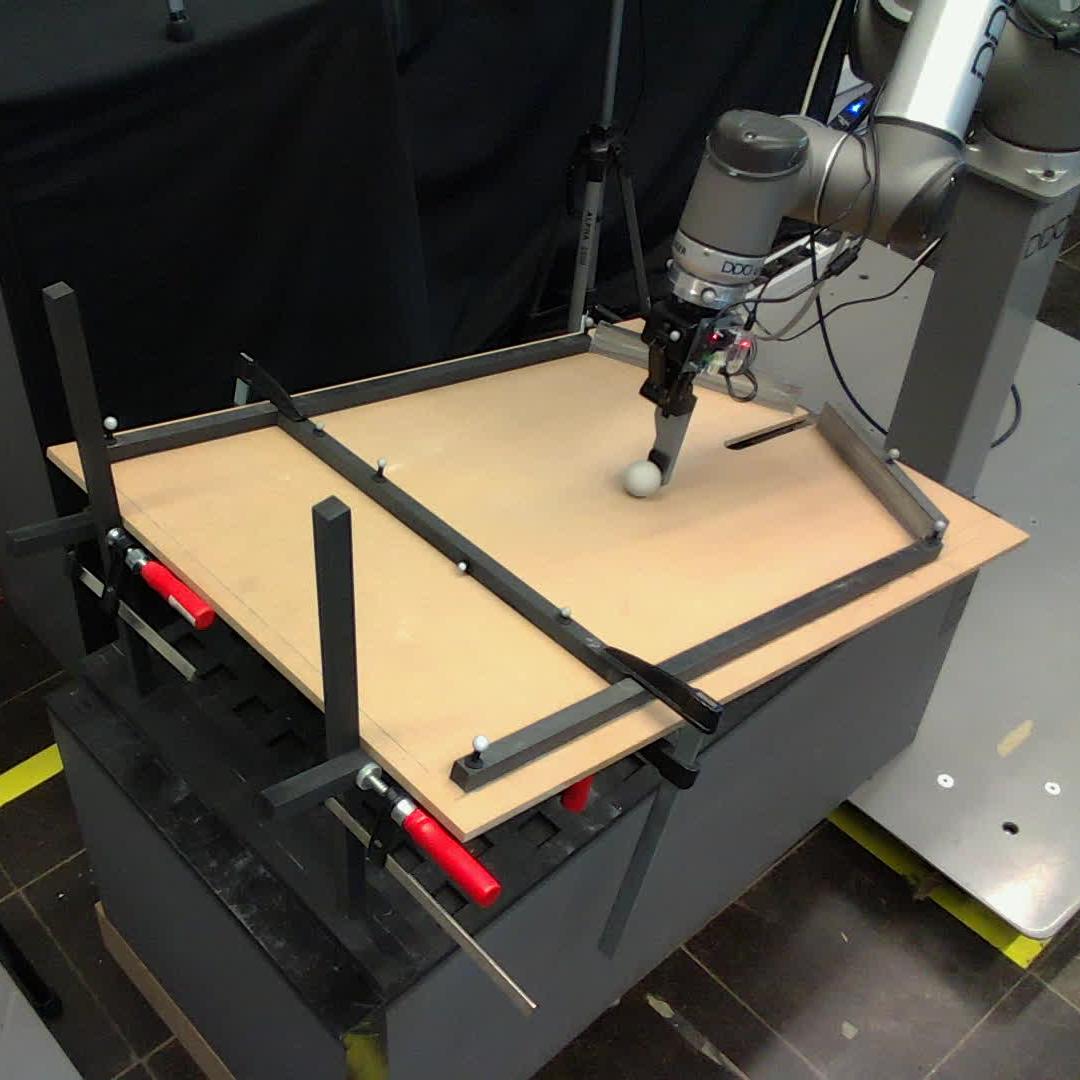}
    \end{subfigure}
    \caption{Our \textit{active exploration} strategy evaluated on a challenging
    \textit{\balltaskname} task in simulation (left) and on the real robot (right).
    The agent needs to learn the dynamics of the task and identify a sparse reward model in order to bring the ball to a target location.
    The robot learns to solve the task online in the real setting from scratch.}
    \label{fig:front_fig}
\end{figure}

A common view in cognitive science is that the evolution of dexterous manipulation capabilities was one of the major driving factors in the development of the human mind~\cite{macdougall1905significance} and the success of humankind in general~\cite{young2003evolution}.
Performing manipulation is cognitively highly demanding, forcing the agent to reason not only about the impact of its actions on itself, but also on the environment.
This inherent complexity leaves autonomous robotic manipulation a largely unsolved problem, despite significant advances in robotics and machine learning in the last decades~\cite{kroemer2021review}.

One of the central challenges of manipulation is the uncertainty about the environment.
When an object is manipulated, its physical properties are rarely known in advance.
Instead, they must be inferred from observations and touch.
To deal with such inference problems effectively, humans have developed various active haptic exploration strategies~\cite{lacreuse1997manual,turvey1995dynamic}.

Prominent approaches in robotic manipulation span from motion planning methods~\cite{li1989motion} to imitation~\cite{fang2019survey} and reinforcement learning~\cite{kalashnikov2018scalable}.
Motion planning usually suffers from ill-defined task descriptions that combined with uninformed prior trajectory distribution lead to suboptimal behaviors.
Learning-based methods, on the other hand, overfit single solutions, and collapse to low-entropy behaviors that fail to generalize to unseen variations of the same task.
We believe that for robots to reach human-level manipulation skills, they must \textit{actively explore} and adapt to new instances of a task.

We define \textit{active exploration} as the directed search of the agent, during the learning process, for unvisited state-action pairs that would maximize the agent's performance.
In this work, we draw inspiration from the \gls{ai}~\cite{parr2022active} field of studies to propose an active exploration framework for model-based reinforcement learning of challenging robotic manipulation tasks.
In our problem formulation, we consider a fully observable environment with unknown world model whose dynamics we need to learn.

Unlike other approaches, that either couple model-free and model-based reinforcement learning to achieve better model learning through costly maximum entropy exploration~\cite{janner2019trust, morgan2021model,hafner2019dream}, or introduce intrinsic signals related to the learned model variance to promote exploration~\cite{pathak2017curiosity,pathak2019self}, we take an information-theoretic perspective, under the umbrella of \gls{ai}.
Starting from a common framework for model learning, namely, using an ensemble of neural networks that allow the estimation of epistemic uncertainty, we propose to use an information-seeking \gls{mpc}, that systematically explores in environments with sparse rewards.

Our information-seeking \gls{mpc} tries to select actions that maximize the agent's information gain, balancing between highly exploratory actions when the dynamics model is unknown and the task performance when the agent has better confidence about its knowledge of the world model.
We provide a thorough theoretical analysis regarding the implementation of our active exploration framework in model-based reinforcement learning.
We evaluate our algorithmic contribution in simulated tasks of increased difficulty, where a 6 degrees of freedom (dof) robotic arm aims to solve a task where it has to push a ball on a tilted table to reach a goal position (\cref{fig:front_fig}).
This task, though seemingly simple, introduces many challenges, as the robot has to balance the ball at the tip of its end-effector, and push the ball over a tilted table to reach an unknown for the agent goal.
Moreover, we provide proof of concept results on a real robotic system that learns online, and we provide our insights regarding real-world robot learning with challenging dynamics.

To summarize our contributions in this work,
\begin{itemize}
    \item we derive a novel algorithm for information-seeking model-based reinforcement learning,
    \item we investigate different measures of curiosity and information seeking strategies that can promote exploration for better dynamics model learning in simulated ball-pushing tasks with different difficulty levels, and
    \item we demonstrate a real-world execution of our actively exploring model-based controller for 7-dof manipulator learning to push a ball over a tilted table.
\end{itemize}

%% file: s1_related_work.tex
\section{Related Work}\label{sec:related}

This field of research attempts to solve \gls{rl} problems by predicting actions as an a posteriori estimate of trajectory rollouts given a prior distribution of actions ~\cite{deisenroth2011pilco}. 
Model-based \gls{rl} learns the transition dynamics of the Markov decision process (MDP) and solves an optimal control problem, usually employing \gls{mpc}. 
On the one hand, model-based \gls{rl} is a promising way for learning reactive robot control strategies, benefiting from the integrated planner that can additionally integrate constraints about the problem. 
Applications of model-based \gls{rl} can be found for robot in-hand manipulation \cite{morgan2021model}, human-robot interaction \cite{roveda2020model,chalvatzaki2019learn} and robot manipulation skill learning \cite{pong2018temporal,chebotar2017combining}.

On the other hand, learning the dynamics of the environment, especially in robotics, is a very challenging problem. 
Real-world dynamics, particularly the dynamics of complex manipulation tasks, like the task of interest in this paper, are characterized by multimodality that function approximators cannot easily capture. 
A prominent approach in the last years for model-based \gls{rl} is the use of an ensemble of neural networks that can learn from different instances of the collected dataset to capture the epistemic and aleatoric uncertainty of the probabilistic model, thus, attempting to learn more accurate models~\cite{chua2018deep}. 
These probabilistic ensembles are coupled with trajectory sampling (PETS)~\cite{chua2018deep} via the cross-entropy method to estimate the best actions to apply to the environment. 
However, PETS suffers from the local exploration in the model rollouts that do not drive the agent to unknown states in the environment, therefore, it learns suboptimal dynamics models. 
Following methods attempt to couple model-free exploration with model learning, as in model-based policy optimization (MBPO)~\cite{janner2019trust}, but the purpose is to accelerate policy learning by utilizing model-based approximate samples.
While this method is faster in terms of convergence compared to its pure model-free competitor \gls{sac}, it still needs an impractical amount of samples to learn a good control policy.
MoPAC~\cite{morgan2021model} improved over MBPO by employing model-predictive rollouts in the approximate MDP that is learned through the model, incentivizing the agent to explore areas of the state-space where model predictions are inefficient.
A parallel line of works aims to learn world models from images, i.e., for observations, using a variational autoencoder to encode the image features and a probabilistic model to encode the dynamics with \gls{cem} for planning, like PlaNet ~\cite{hafner2019learning}.
Unlike PETS and PlaNet, which greedily select the actions that the \gls{cem} predicts to yield the highest reward, we propose a novel framework for model-based \gls{rl} using an information-seeking objective in our \gls{mpc}, that balances exploration-exploitation during learning, by promoting maximization of the information again when the model is still suboptimal and optimizes for the end-task once the agent is confident about its knowledge about the world dynamics. 

Exploration for efficient model learning is an open research topic, with intrinsic motivation being a well-known method for exploration towards model learning~\cite{singh2010intrinsically, hester2017intrinsically}.
However, intrinsic rewards usually rely on hand-crafted metrics of learning progress, making them difficult to apply in a wide range of tasks~\cite{lopes2012exploration}.
In the era of the deep probabilistic ensembles, a self-supervised way of encouraging exploration of states that will improve model learning are methods based on ensemble disagreement, where the variance in the predictions of the different ensemble models is used as intrinsic reward for a model-free policy~\cite{pathak2019self,shyam2019model,sekar2020planning}.
This policy, thus, learns to collect data in areas of the state-space where the disagreement between the predictions of the ensemble models is higher.
However, advantages of these methods compared to MBPO, MoPAC and other model-based methods without explicit exploration bonus~\cite{hafner2019dream} are not well-established.
Even if disagreement promotes exploration in early stages, it is more beneficial to vision-based \gls{rl} settings, where the variance between the ensembles can be high due to the high image reconstruction errors, but its benefit to trajectory-based MDPs is incremental, as the ensembles are bounded by the statistics of the marginal state distribution of the dataset.
Other methods for inducing curiosity in \gls{rl} rely on prediction error~\cite{pathak2017curiosity}, epistemic uncertainty~\cite{Bechtle2019Apr} or state visitation counts~\cite{Ecoffet2019Jan} as reward signals during roll-outs.
Contrary to these methods that introduce heuristic intrinsic rewards or rely on a model-free policy for model learning, our method employs a principled information-theoretic approach that can be traced back to early works on Bayesian experiment design~\cite{lindley1956measure}.
Specifically, we utilize \gls{mpc} to trade off between actions that yield a high expected extrinsic reward and actions that maximize the expected information gain of the observed states for our model in real time.
The use of an expected information gain term inside \gls{mpc} allows our agent to plan even beyond the space of states it has visited so far to obtain observations that it expects to be beneficial for model learning.

%% file: s2_method.tex
\section{Active Exploration for Model Learning}

We assume that the environment is fully observable, governed by unknown dynamics $\PP{\vs_\tau}[\vs_{\tau - 1}, \va_\tau]$, and provides the agent with an a-priori unknown reward $\PP{\vr_\tau}[\vs_{\tau}, \va_\tau]$ in every time step.
Here, $\vs_{\tau} \in \R^{N_\vs}$ denotes the state of the environment at time $\tau$, $\va_\tau \in \R^{N_\va}$ the action the agent can take, and $\vr_\tau \in \R$ the reward it is receiving.
The agent's objective is the maximization of the cumulative reward over a fixed time horizon of $T$ discrete time steps, that is
\begin{align}
    \max_{\pi} \quad \Ex{\PP{\mathbf{\vr}_{1:T}}[\pi]}{\sum_{\tau = 1}^T \vr_{\tau}}
\end{align}
where $\pr{\pi}{\va_\tau}[\vs_{\tau - 1}]$ is the agent's policy.

Since the real dynamics and reward distributions are unknown, the agent maintains approximations $\p{\vs_\tau}[\va_\tau, \vs_{\tau -1}, \theta]$ and $\p{r_\tau}[\vs_{\tau}, \va_\tau, \theta]$ of them, where $\theta$ are the model parameters.
Hence, the agent assumes the following generative model of the environment:
\begin{align}
    \label{eq:model_mdp}
    \begin{split}
        &\p{\mathbf{\vs}_{0:T}, \mathbf{\va}_{1:T}, \mathbf{\vr}_{1:T}, \theta}
        =
        \p[k]{\vs_0}
        \p{\theta}
        \prod_{\tau = 1}^T
        \Big(
        \p{\vr_\tau}[\vs_{\tau}, \va_\tau, \theta] \rhookswarrow
        \\
        &\qquad\p{\vs_\tau}[\vs_{\tau - 1}, \va_\tau, \theta]
        \pr{\pi}{\va_\tau}[\vs_{\tau - 1}] \Big)
    \end{split}
\end{align}
where $\p[k]{\theta}$ is the agent's belief over the correct model parameters in episode $k$.

Instead of greedily optimizing the expected reward directly, we propose to optimize the sum of the expected reward and an intrinsic term that encourages the agent to make observations informative w.r.t.\ its model.
Hence, at time $t$, we choose $\pi$ such that it optimizes the problem
\begin{align}
    \max_{\pi} \;
    \Ex{\p{\mathbf{\vr}_{t+1:t+H}}[\pi]}{\sum_{\tau = t + 1}^{t + H} \vr_{\tau}}
    +
    \beta \text{I} \left( \pi, \vs_t \right)
    \label{eq:obj}
\end{align}
where $H \in \mathbb{N}$ is the planning horizon, $\beta \in \R$ a weighting factor, and $\text{I} \left( \pi, \vs_t \right)$ the intrinsic term.

In this work, we choose the intrinsic term to be the information gain between the model parameters and the expected states and rewards:
\begin{align}
    \text{I} \left( \pi, \vs_t \right)
    &\coloneqq
    \text{MI} \left( \left(\mathbf{\vs}, \mathbf{\vr} \right), \theta \mid \pi, \vs_t \right)
    \\
    &=
    \Ex{\p{\mathbf{\vs}, \mathbf{\vr}}[\pi, \vs_t]}{
        \kl{\p{\theta}[\mathbf{\vs}, \mathbf{\vr}, \pi, \vs_t]}{\p{\theta}}
    }
\end{align}
where $\mathbf{\vs} \coloneqq \mathbf{\vs}_{t+1:t+H}$ and $\mathbf{\vr} \coloneqq \mathbf{\vr}_{t+1:t+H}$ is the sequence of states and rewards up to the planning horizon.
Note that in the literature, the expected information gain is also known as \glsfirst{mi}, which is why we denoted it accordingly in the equation above.

The expected information gain can be seen as a measure of how much the agent expects to learn about the environment by following policy $\pi$ in state $s_t$.
Specifically, this measure becomes maximal if the agent expects to make observations that will change its belief about the correct choice of model parameters drastically.
Hence, an agent maximizing this term will be curious about its environment and explore it systematically, even in the total absence of extrinsic reward.
In combination with the expected reward, we obtain an agent that is acting both information-seeking and goal-directed, with the trade-off being explicitly controlled by the weighting factor $\beta$.

The optimization of the objective can now be performed by any planner that is capable of handling continuous action spaces.
In this work, we use a variant of the \glsxtrlong{cem} to find an open loop sequence of actions $\mathbf{\va}_{t+1:T}$ that maximizes the objective.

\subsection{Approximation of the planning objective}
A major challenge in computing the joint objective is that neither the expected reward nor the intrinsic term can be computed in closed form.
While the expected reward can straightforwardly be approximated via \gls{mc}~\cite{chua2018deep,hafner2019learning}, the intrinsic term is known to be notoriously difficult to compute~\cite{agakov2004algorithm,foster2019variational,belghazi2018mine,poole2019variational}.
Thus, instead of maximizing \gls{mi} directly, many methods maximize a variational lower bound of it~\cite{poole2019variational}.
However, due to the high-dimensional nature of $\theta$, these approaches are too expensive to be executed during planning in real time.

Hence, instead we propose to use a \gls{nmc} estimator that reuses samples from the outer estimator in the inner estimator:
\begin{align}
    \label{eq:meth_nmc_est_reuse}
    &\text{MI} \left( \left(\mathbf{\vs}, \mathbf{\vr} \right), \theta \mid \pi, \vs_t \right)
    \\
    &=
    \Ex{\p{\theta}}{
        \Ex{\p{\mathbf{\vs}, \mathbf{\vr}}[\pi, \vs_t, \theta]}{
            \ln \p{\mathbf{\vs}, \mathbf{\vr}}[\pi, \vs_t, \theta]
            - \ln \p{\mathbf{\vs}, \mathbf{\vr}}
        }
    }
    \\
    &\approx
    \underbrace{
        \frac{1}{n}
        \sum_{i=1}^{n}
        ( \ln \p{\vs^{i}, \vr^i}[\theta^{i}]
        -
        \ln
        \underbrace{
            \frac{1}{n}
            \sum_{\substack{k=1\\k\neq i}}^{n}
            \p{\vs^{i}, \vr^i}[\theta^{k}]
        }_{\text{inner estimator}}
        )}_{\text{outer estimator}}
\end{align}
where
\begin{equation}
    \theta^{i} \sim \p{\theta}, \quad \left(\mathbf{\vs}^i, \mathbf{\vr}^i \right) \sim \p{\mathbf{\vs}, \mathbf{\vr}}[\theta^{i}], \quad \forall i \in \left\{ 1, \dots, n \right\}.
\end{equation}
Although using the same samples $\theta_1, \dots, \theta_n$ in the inner estimator as in the outer estimator violates the \gls{iid} assumption, we found this reuse of samples can substantially increase the sample efficiency.

\subsection{Choice of model}
We approximate the dynamics and the reward models with Gaussian distributions where the mean and covariance are given by neural networks with weights $\theta$
\begin{align}
    \label{eq:model_def}
    \begin{split}
        \p{\vs_\tau}[\va_\tau, \vs_{\tau -1}, \theta]
        &\coloneqq
        \mathcal{N} \left( \vs_\tau \mid \mu^{\vs}_\theta \left( \vs_{\tau - 1}, \va_\tau \right), \Sigma^{\vs}_\theta \left( \vs_{\tau - 1}, \va_\tau \right) \right)
        \\
        \p{r_\tau}[\vs_{\tau}, \va_\tau, \theta]
        &\coloneqq
        \mathcal{N} \left( r_\tau \mid \mu^r_\theta \left( \vs_\tau, \va_\tau \right), \sigma^r_\theta \left( \vs_\tau, \va_\tau \right) \right).
    \end{split}
\end{align}

There are multiple options for representing distributions over neural network parameters.
Common choices are particle-based representations~\cite{opitz1999popular,tschantz2020reinforcement}, Gaussian distributions with diagonal covariance matrix~\cite{lampinen2001bayesian} or a combination of both~\cite{jospin2020hands}.
Since our approximation of the planning objective only requires samples of $\theta$, we choose to represent $\p[k]{\theta}$ by a set of particles $\theta_1, \dots, \theta_n$, making our model a neural network ensemble.

\subsection{Lautum Information}
\begin{figure}[t]
    \centering
    \begin{tikzpicture}
        \begin{axis}
            [
            title={\small Cosine similarity over sample count},
            align=center,
            legend style={nodes={scale=0.8, transform shape}},
            grid=both,
            width=\linewidth,
            height=4.5cm,
            reverse legend,
            legend pos=south east,
            legend cell align={left},
            ymin=0.8,
            xlabel = \small{Total number of samples},
            ylabel = \small{Cosine similarity}]
            \plotWithStddev{data/li_mi/mi_cos_sim.csv}{pltRed}{1}{\gls{mi}}{sample_count}{mean}{std}
            \plotWithStddev{data/li_mi/li_cos_sim.csv}{pltBlue}{1}{\gls{li}}{sample_count}{mean}{std}
        \end{axis}
    \end{tikzpicture}
    \caption{
        Comparison of the cosine similarity between the sample-reusing \gls{nmc} approximations of \gls{mi} and \gls{li} to their respective exact values.
        For each number of samples, we conducted 1000 experiments on randomly generated discrete generative models $\p{\vs, \theta}[\pi]$.
        To assess the approximation quality independently of scale, we compute the cosine similarity between the approximated and exact information vectors $\left( \text{I} \left( \vs, \theta \mid \pi_1 \right), \dots, \left( \vs, \theta \mid \pi_m \right) \right)^T$, where I is either \gls{mi} or \gls{li} and $m$ is the number of policies in the model.
        \textit{Number of samples} refers to the sum of $\vs$ and $\theta$ samples.
        Note that the optimal value possible under cosine similarity is 1.
    }
    \label{fig:li_exp1_fig}
    \vspace{-1em}
\end{figure}
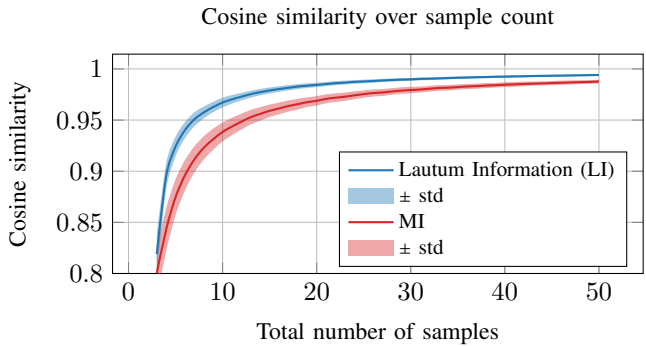

An alternative to our choice of the intrinsic term is to use the reverse \gls{kl} divergence, yielding the \gls{li}~\cite{palomar2008lautum}, which in our case is defined as
\begin{align}
    \text{I} \left( \pi, \vs_t \right)
    &\coloneqq
    \text{LI} \left( \left(\mathbf{\vs}, \mathbf{\vr} \right), \theta \mid \pi, \vs_t \right)
    \\
    &=
    \Ex{\p{\mathbf{\vs}, \mathbf{\vr}}[\pi, \vs_t]}{
        \kl{\p{\theta}}{\p{\theta}[\mathbf{\vs}, \mathbf{\vr}, \pi, \vs_t]}
    }.
\end{align}
\gls{li}, similar to \gls{mi}, measures how much information we are expected to gain about $\theta$ by observing $\left(\mathbf{\vs}, \mathbf{\vr} \right)$. However, it does it in a different way and leads to a different result, similar to how reverse KL leads to mode-seeking behavior and the forward KL to the moment matching behavior.

To gain an intuition about the difference between \gls{mi} and \gls{li}, it is useful to consider which policy $\pi$ maximizes each of them.
For \gls{li}, the prior $\p{\theta}$ is in the numerator in the \gls{kl} divergence term,
therefore \gls{li} becomes maximal for policies that are expected to produce observations which make a-priori likely $\theta$ have a low probability in the posterior.
In a sense, \gls{li} encourages the agent to seek out observations that disprove the optimality of those $\theta$ that the agent assigned a high prior probability to.
Analogously, \gls{mi} encourages the agent to make observations that cause $\theta$ with a low prior probability to have a high posterior probability.

One advantage of \gls{li} is the independence of $\left(\mathbf{\vs}, \mathbf{\vr} \right)$ and $\theta$ in the outer expectations
\begin{equation}
    \text{LI} =
    \Ex{\p{\theta}}{
        \Ex{\p{\mathbf{\vs}, \mathbf{\vr}}[\pi, \vs_t]}{
            \ln \p{\mathbf{\vs}, \mathbf{\vr}} - \ln \p{\mathbf{\vs}, \mathbf{\vr}}[\pi, \vs_t, \theta]
        }
    }
\end{equation}
which allows for a more efficient reuse of samples.
Contrary to the \gls{mi} approximation, when approximating the inner expectation $\Ex{\p{\mathbf{\vs}, \mathbf{\vr}}[\pi, \vs_t]}{.}$, we can reuse the same samples $\left(\mathbf{\vs}^i, \mathbf{\vr}^i \right)$ for all samples $\theta^j$ from the outer expectation.
Hence, the resulting \gls{nmc} approximation of \gls{li} is given as
\begin{equation}
    \label{eq:meth_li_reuse_nmc}
    \text{LI}
    \approx
    \frac{1}{n}
    \sum_{i=1}^{n}
    \ln
    \left(
    \frac{1}{n}
    \sum_{\substack{k=1\\k\neq i}}^{n}
    \p{\mathbf{\vs}^i, \mathbf{\vr}^i}[\theta^{k}]
    \right)
    -
    \frac{1}{n}
    \sum_{\substack{j=1\\j\neq i}}^{n}
    \ln \p{\mathbf{\vs}^i, \mathbf{\vr}^i}[\theta^{j}].
\end{equation}
An empirical comparison of the stochastic estimators of \gls{mi} and  \gls{li} is shown in~\cref{fig:li_exp1_fig},
which suggest a clear advantage in sample efficiency of the \gls{li} approximation in comparison to the \gls{mi} approximation.
Theoretical analysis of these estimators may be of interest for future work.
The practical influence of the choice of the information gain term is evaluated in~\cref{sec:exp}.
To our knowledge, \gls{li} has not been used in the context of exploration before.

Further implementation details as well as the link to our code can be found on our project page:\\\url{https://sites.google.com/view/aerm}.

%% file: s3_experiments.tex
\section{Experimental Results}
\label{sec:exp}

\begin{figure}[t]
    \centering
    \begin{subfigure}[t]{0.32\linewidth}
        \centering
        \includegraphics[width=\linewidth]{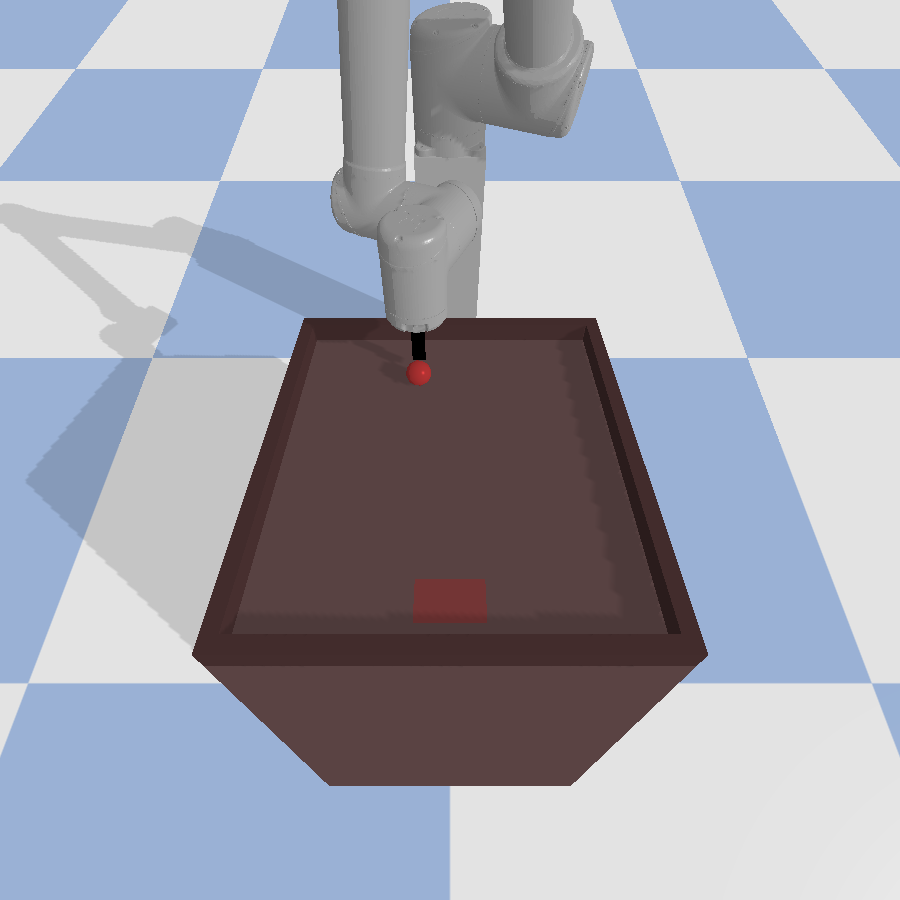}
        \caption{\textit{\balltaskname}}
    \end{subfigure}
    \begin{subfigure}[t]{0.32\linewidth}
        \centering
        \includegraphics[width=\linewidth]{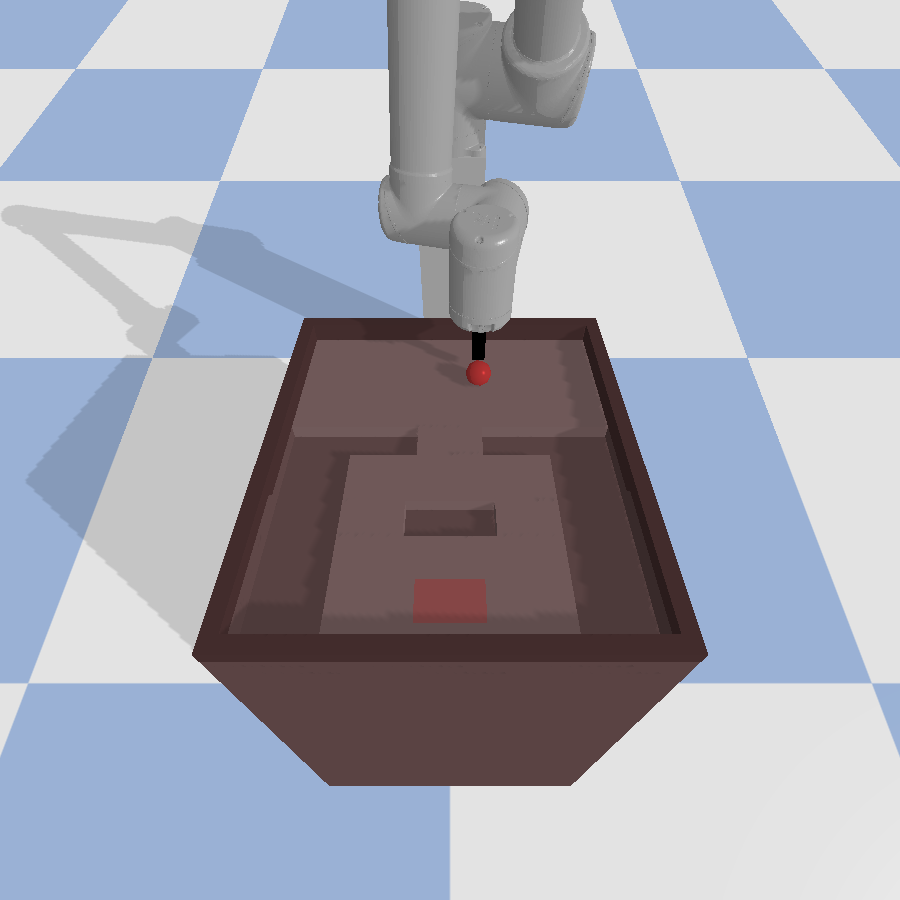}
        \caption{\textit{\ballhtaskname}}
    \end{subfigure}
    \begin{subfigure}[t]{0.32\linewidth}
        \centering
        \includegraphics[width=\linewidth]{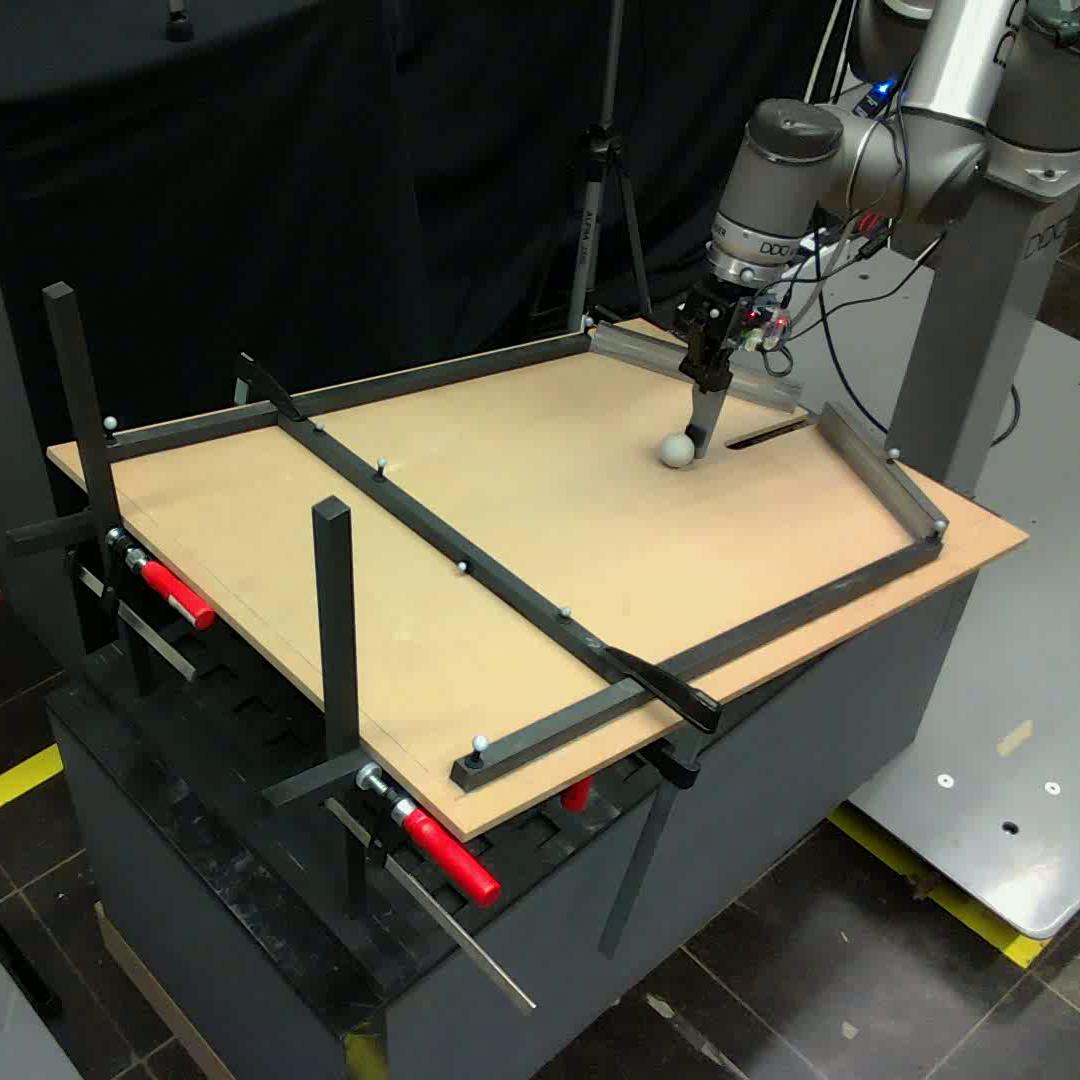}
        \caption{\textit{\ballrtaskname}}
        \label{fig:tasks_real}
    \end{subfigure}
    \caption{
        Visualization of the three environment configurations we test our methods on.
        The target zone (marked in red in the simulated environments) is always at the top center of the table and its location is not observed by the agent. 
    }
    \label{fig:tasks}
\end{figure}

\begin{figure*}[t]
    \centering
    \begin{subfigure}[t]{0.32\linewidth}
        \centering
        \begin{tikzpicture}
            \begin{axis}[rewplot, ylabel = {\footnotesize Cumulative reward}, legend style={at={(0.98,0.44)},anchor=east}]
                \plotstdcom{data/ball_t30r1f05/baselines/mbpo_reward_test.csv}{pltBrown}{1}{MBPO}{step}{mean}[std][1][50]
                \plotstdcom{data/ball_t30r1f02/baselines/sac_reward_test.csv}{pltPurple}{1}{SAC}{step}{mean}[std][1][50]
                \plotstdcom{data/ball_t30r1f02/ours/no_intrinsic_reward_test.csv}{pltOrange}{1}{PETS}{step}{mean}[std][1][50]
                \plotstdcom{data/ball_t30r1f02/ours/li_reward_test.csv}{pltRed}{1}{\gls{li}}{step}{mean}[std][1][50]
                \plotstdcom{data/ball_t30r1f02/ours/mi_reward_test.csv}{pltBlue}{1}{\gls{mi}}{step}{mean}[std][1][50]
            \end{axis}
        \end{tikzpicture}
        \caption{\textit{\balltaskname}}
    \end{subfigure}
    \begin{subfigure}[t]{0.32\linewidth}
        \centering
        \begin{tikzpicture}
            \begin{axis}[rewplot]
                \plotstdcom{data/ball_t30r1f05h01/baselines/mbpo_reward_test.csv}{pltBrown}{1}{MBPO}{step}{mean}[std][1][50]
                \plotstdcom{data/ball_t30r1f05h01/baselines/sac_reward_test.csv}{pltPurple}{1}{SAC}{step}{mean}[std][1][50]
                \plotstdcom{data/ball_t30r1f05h01/ours/no_intrinsic_reward_test.csv}{pltOrange}{1}{PETS}{step}{mean}[std][1][50]
                \plotstdcom{data/ball_t30r1f05h01/ours/li_reward_test.csv}{pltRed}{1}{\gls{li}}{step}{mean}[std][1][50]
                \plotstdcom{data/ball_t30r1f05h01/ours/mi_reward_test.csv}{pltBlue}{1}{\gls{mi}}{step}{mean}[std][1][50]
                \legend{};
            \end{axis}
        \end{tikzpicture}
        \caption{\textit{\ballhtaskname}}
    \end{subfigure}
    \begin{subfigure}[t]{0.32\linewidth}
        \centering
        \begin{tikzpicture}
            \begin{axis}[rewplot]
                \plotstdcom{data/ball-real_ca0r0/ours/mi_reward_test.csv}{pltBlue}{1}{\gls{mi}}{step}{mean_smooth}[std_smooth][1][30]
                \legend{};
                \pgfplotstableread[col sep=comma]{data/ball-real_ca0r0/ours/mi_no_var_reward_test.csv}\datatable
                \addplot+[color=pltBlue, smooth, dotted] table [x expr=30 * \thisrow{step}, y expr=\thisrow{mean_smooth}]{\datatable};
                \addlegendentry{no learn var.}
            \end{axis}
        \end{tikzpicture}
        \caption{\textit{\ballrtaskname}}
        \label{fig:exp_ball_res_real}
    \end{subfigure}
    \caption{
        Cumulative reward in three different experiments for both versions of our agent (\gls{mi} and \gls{li}) in comparison to three baselines: PETS~\cite{chua2018deep}, SAC~\cite{haarnoja2018soft}, and MBPO~\cite{janner2019trust}.
        The graphs display the evaluation reward, which is obtained by rolling out the learned model or policy without adding intrinsic reward or action noise.
        In both simulated tasks (\textit{\balltaskname} and \textit{\ballhtaskname}), all baselines failed  to discover the reward at the top of the table and converged to local minima.
        Our configurations found the reward consistently within 1,500,000 steps except for one of the five \gls{li} runs on the \textit{\ballhtaskname} task.
        On the real robot we evaluate only the \gls{mi} configuration, since each experiment of 150,000 steps corresponds to approximately 21 hours of training time.
        The dashed blue line in (\subref{fig:exp_ball_res_real}) indicates a continuation of the experiment where we did not start to learn transition variances after 60,000 steps.
        Note that the episode length of the simulation experiments is 50 steps, while we only use 30 steps for the real-world experiments to speed up the training. 
    }
    \label{fig:exp_ball_res}
    \vspace{-1em}
\end{figure*}
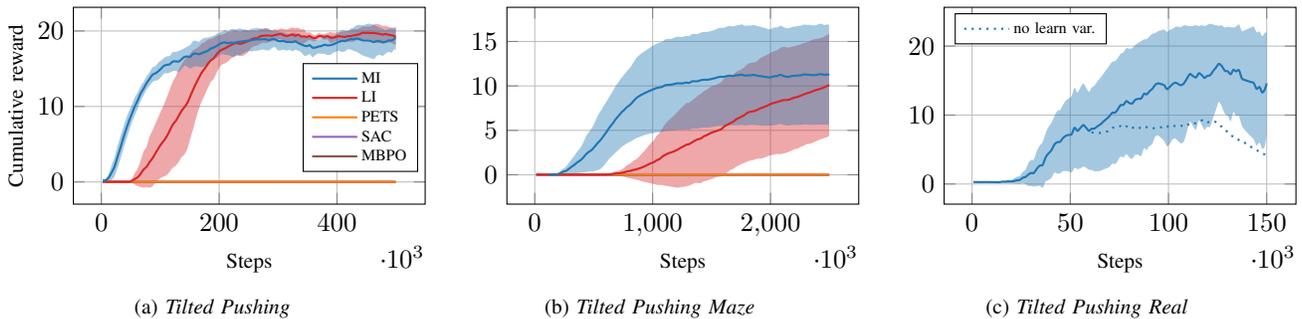

A central feature that sets our method apart from other purely model-based approaches~\cite{hafner2019learning,chua2018deep} is the intrinsic term, that explicitly drives the agent to explore its environment in a systematic manner.
Since to our knowledge, there exists no well established benchmark for hard-to-explore continuous-control robotic manipulation tasks yet, we designed two such tasks in simulation, \textit{\balltaskname} and \textit{\ballhtaskname}), and one in the real world, \textit{\ballrtaskname} (see \cref{fig:tasks}).
In all tasks, the agent has to push a ball up a tilted table into a target zone to receive a reward.
The agent can move the gripper in a plane parallel to the table and, in the simulated tasks, also rotate the black end-effector around the Z-axis (Z-axis being orthogonal to the brown table and pointing up).
In the real world experiments we disabled end-effector rotation to make the task complexity and thereby also the training time more manageable.
As input, the agent receives the 2D positions and velocities of both the gripper and the ball, and the angular position and velocity of the end-effector. 
For the real robot training we rely on motion capture for measuring the state of the environment.
To test the limits of our methods, in the \textit{\ballhtaskname} task we add holes to the table, that irrecoverably trap the ball if it falls in.

There are two aspects that make these tasks particularly challenging.
First, the reward is sparse, meaning that the only way the agent can learn about the reward at the top of the table is by moving the ball there and exploring it.
The agent receives 1 when pushing the ball in the target goal area, and 0 otherwise, except for a small penalty for large actions.
Second, balancing the ball on the finger and moving it around is non-trivial and requires dexterity, especially given the low control frequency of 4 Hz we operate our agent on.
Note that the robot actions have to be slow, always balancing the ball at the tip of the end-effector, and it doesn't try to achieve the goal by flicking the ball. This makes the combined control and exploration task significantly harder than simple pushing tasks. On top of that, the table is tilted, meaning that the gravity affects the balancing of the ball by the robot, making the dynamics of the task very hard to learn.
Once the agent drops the ball, this cannot be recovered, giving the agent no choice but to wait for the episode to terminate to continue exploring.
Both of these aspects make solving these tasks with conventional, undirected exploration methods like Boltzmann exploration or Gaussian action noise extremely challenging.
Consequently, the agent has to learn to balance the ball without receiving any extrinsic reward, purely driven by its own curiosity.

In both simulated experiments, we compare against three baselines: PETS~\cite{chua2018deep}, SAC~\cite{haarnoja2018soft}, and MBPO~\cite{janner2019trust}.
We chose these three methods, because they are popular examples of each of the three main categories that most modern \gls{rl} algorithms fall into:
SAC is completely model-free, PETS is fully model-based and MBPO is a hybrid approach where the model is used to generate additional data for an underlying model-free agent.
Note that we do not compare to other methods that use other types of intrinsic rewards to induce exploration \cite{pathak2019self}, since, as we described in \cref{sec:related}, those are not pure model-based methods but rely on learning a policy to explore for learning a model.
For a comprehensive overview of exploration methods in \gls{rl}, refer to~\cite{Amin2021Sep}.

For all tasks, our method uses an ensemble of 5 fully-connected neural networks for both dynamics and reward models.
Furthermore, in simulation, we found it beneficial to use a constant standard deviation of 0.001 for both models instead of learning it from the data.

\begin{figure*}[t]
    \centering
    \begin{tabular}{DCCCCC}
        \gls{li} &
        \includegraphics[width=\linewidth]{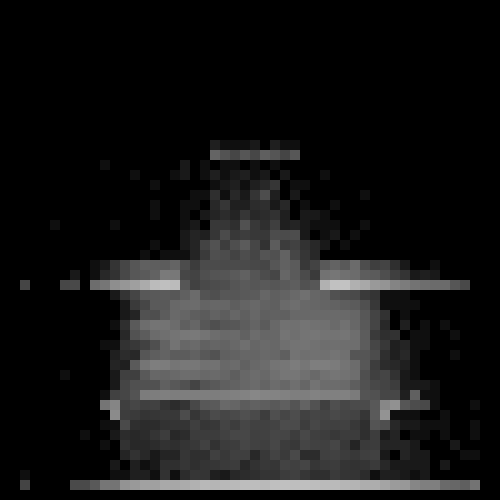} &
        \includegraphics[width=\linewidth]{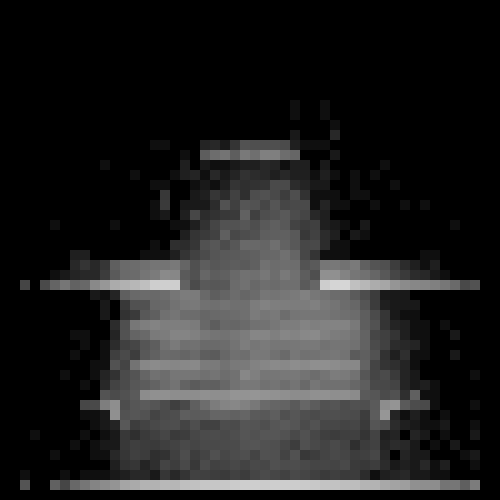} &
        \includegraphics[width=\linewidth]{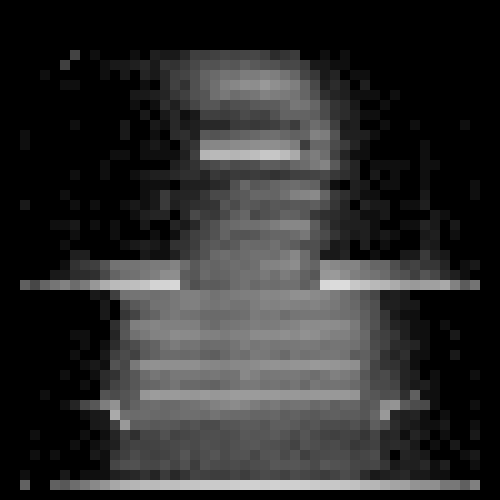} &
        \includegraphics[width=\linewidth]{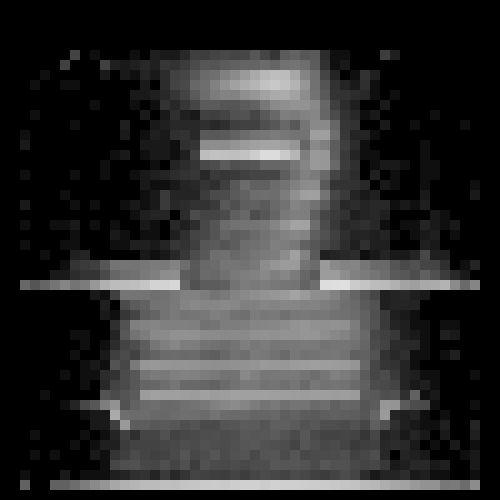} &
        \includegraphics[width=\linewidth]{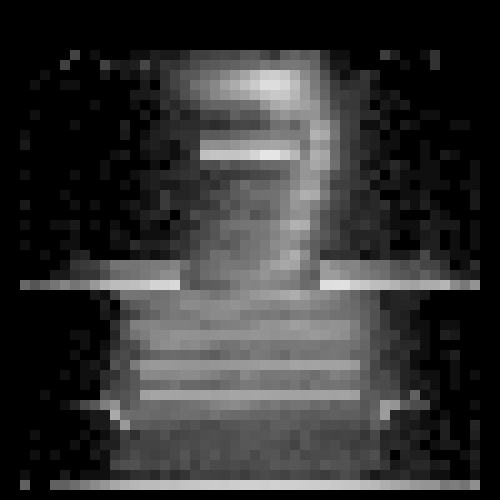}
        \\[1.10cm]
        \gls{mi} &
        \includegraphics[width=\linewidth]{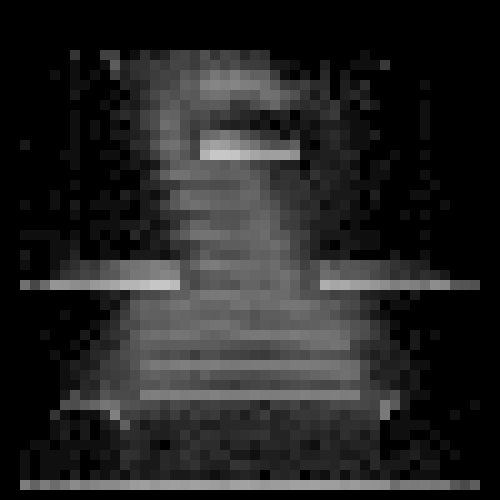} &
        \includegraphics[width=\linewidth]{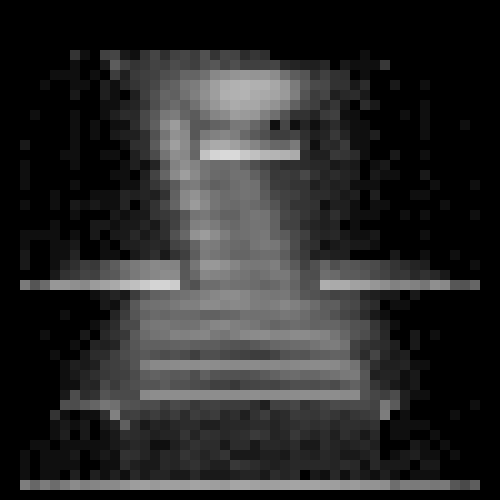} &
        \includegraphics[width=\linewidth]{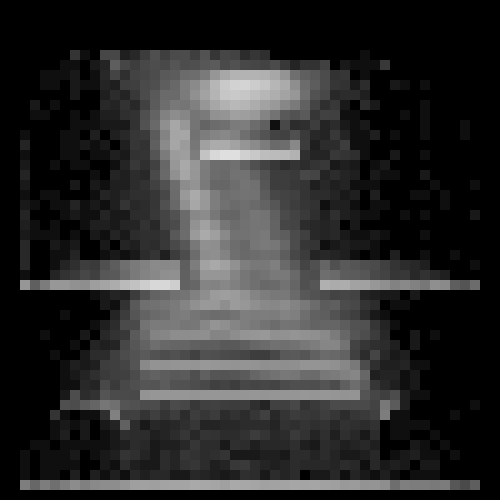} &
        \includegraphics[width=\linewidth]{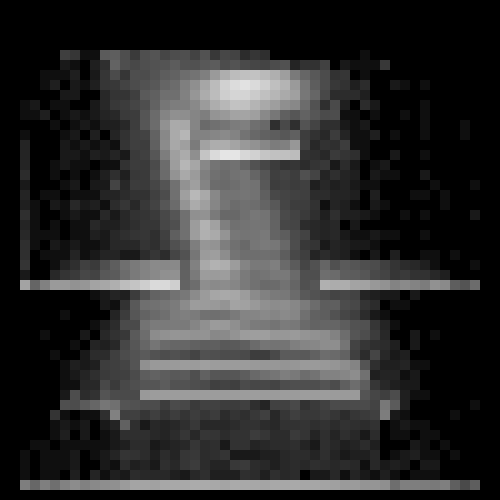} &
        \includegraphics[width=\linewidth]{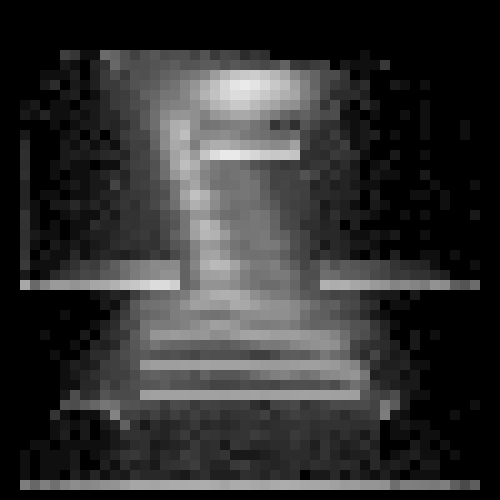}
        \\[1.10cm]
        PETS &
        \includegraphics[width=\linewidth]{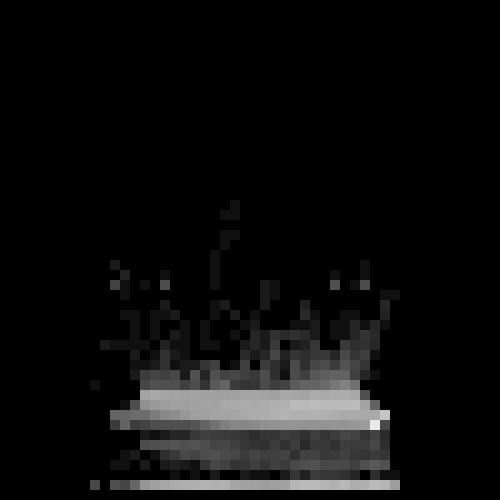} &
        \includegraphics[width=\linewidth]{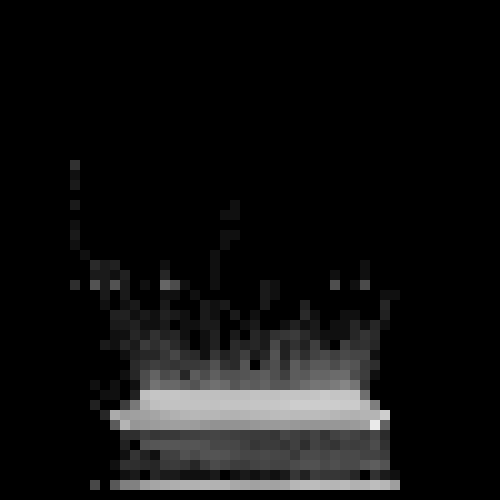} &
        \includegraphics[width=\linewidth]{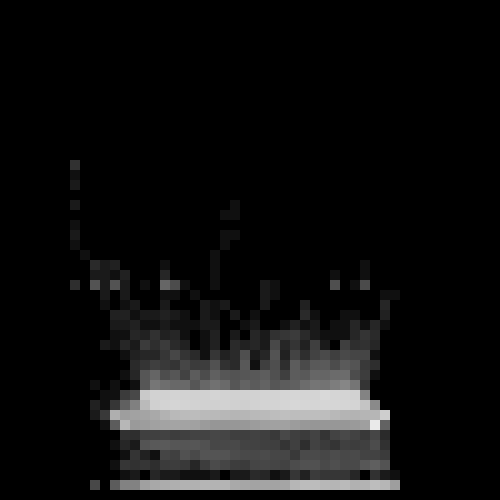} &
        \includegraphics[width=\linewidth]{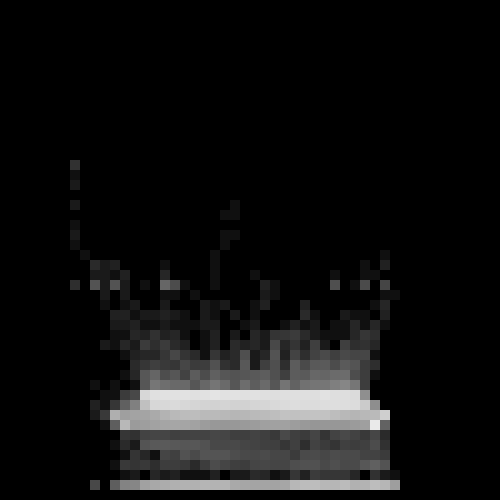} &
        \includegraphics[width=\linewidth]{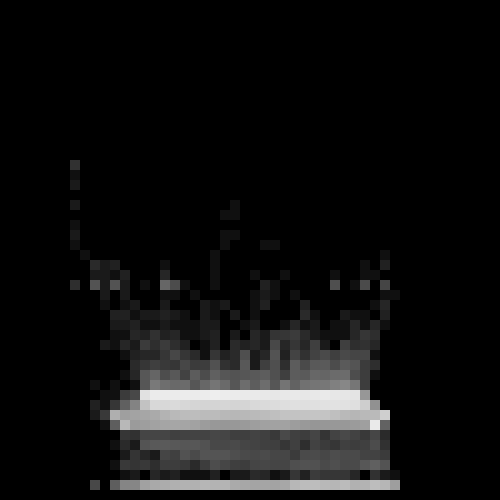}
        \\
        Steps & 150k & 300k & 450k & 600k & 750k
    \end{tabular}\setlength\tabcolsep{6pt}
    \caption{
        Comparison of the states visited by our \gls{mi} and \gls{li} agents and PETS~\cite{chua2018deep} in the \textit{\ballhtaskname}.
        The brightness of each pixel indicates how often the ball has visited the respective point of the table at the given point in the training.
        The coordinate origin is at the bottom of each image, meaning that the images are rotated 180\textdegree{} compared to the top-down view in \cref{fig:tasks}.
        Both the \gls{li} agent and the \gls{mi} agent succeed in solving the task because they achieve a much better coverage than PETS~\cite{chua2018deep}.
    }
    \label{fig:exp_ballh_hist}
\end{figure*}

As shown in \cref{fig:exp_ball_res}, our method is able to solve the \textit{\balltaskname} task consistently with both \gls{mi} and \gls{li} as intrinsic terms.
All baselines fail to find the reward within 300,000 steps and converge to local minima.
Also, the \textit{\ballhtaskname} task is solved consistently within 1,500,000 steps by our method, with only a single agent of the \gls{li} configuration being unable to find the reward in time.
Note, that the holes of \textit{\ballhtaskname} make this environment significantly harder to explore, as the ball has to be maneuvered around two corners in order to reach the target zone.
As can be seen in \cref{fig:exp_ballh_hist}, the reason for the high performance of our method on this task is a much better state space coverage compared to PETS.
While our agents systematically maneuver the ball around the holes in unseen locations, the non-intrinsic agent rarely passes the lower holes and leaves the upper half of the table completely unexplored.
Given that PETS is purely driven by the extrinsic reward, this behavior is not surprising, as these environments initially provide the agents with no direct incentive to learn to balance the ball.
Note that SAC and MBPO with their maximum entropy exploration strategy did not manage to learn the tasks, revealing that hard exploration problems require directed information seeking strategies, as our method does.

To show that our method is principally capable of solving real-world tasks, we re-created the \textit{\balltaskname} in the real world as shown in \cref{fig:tasks_real}.
Applying our approach to the real world, for online training with the robot from scratch, yields a number of additional challenges that the agent has to deal with.
A central challenge we faced is the occurrence of unobserved variables in our environment that violate our Markov assumption.
One such example is that we only observe the position of the ball and not the spin.
Hence, the agent has no way of knowing whether the ball is currently sliding or rolling, yet the future trajectory of the ball may largely depend on this information.
A typical example where information about the ball's spin becomes important is when the robot does some left-right jittering motion.
In this situation, the ball will remain stable as long as it does not start rolling along the finger.
Without knowing the spin, reliably predicting whether the ball will stay on the finger is likely impossible.

We tackled this issue by learning not only the means, but also the variances of the transition model.
In the case of this environment, we found that learned variances cause the planner to avoid jittering motions, as they tend to lead to a high state uncertainty and, thus, a lower expected reward.
However, we also found that learning the variances too early in the training leads to pessimistic behavior, where the agent stops moving at some point in the training despite having found the reward multiple times before. 
So instead of learning the variances directly from the start, we start learning them after 60,000 steps on, when we have collected enough data to predict sensible variances.

\begin{figure*}[t]
    \centering
    \begin{tabular}{DCCCCC}
        5k steps &
        \includegraphics[width=\linewidth]{media/video_frames/ep_00150_s_000150_train/00000.jpg} &
        \includegraphics[width=\linewidth]{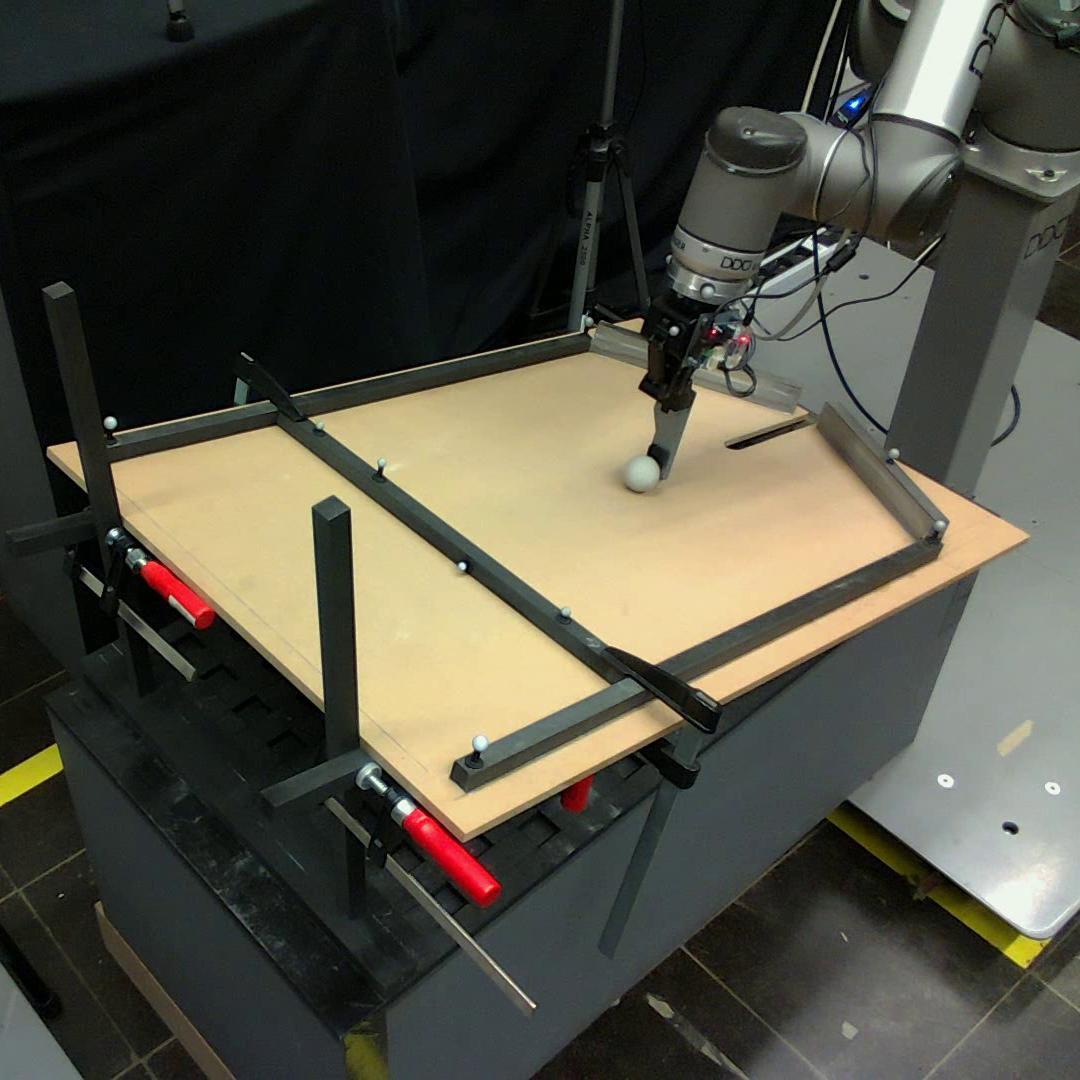} &
        \includegraphics[width=\linewidth]{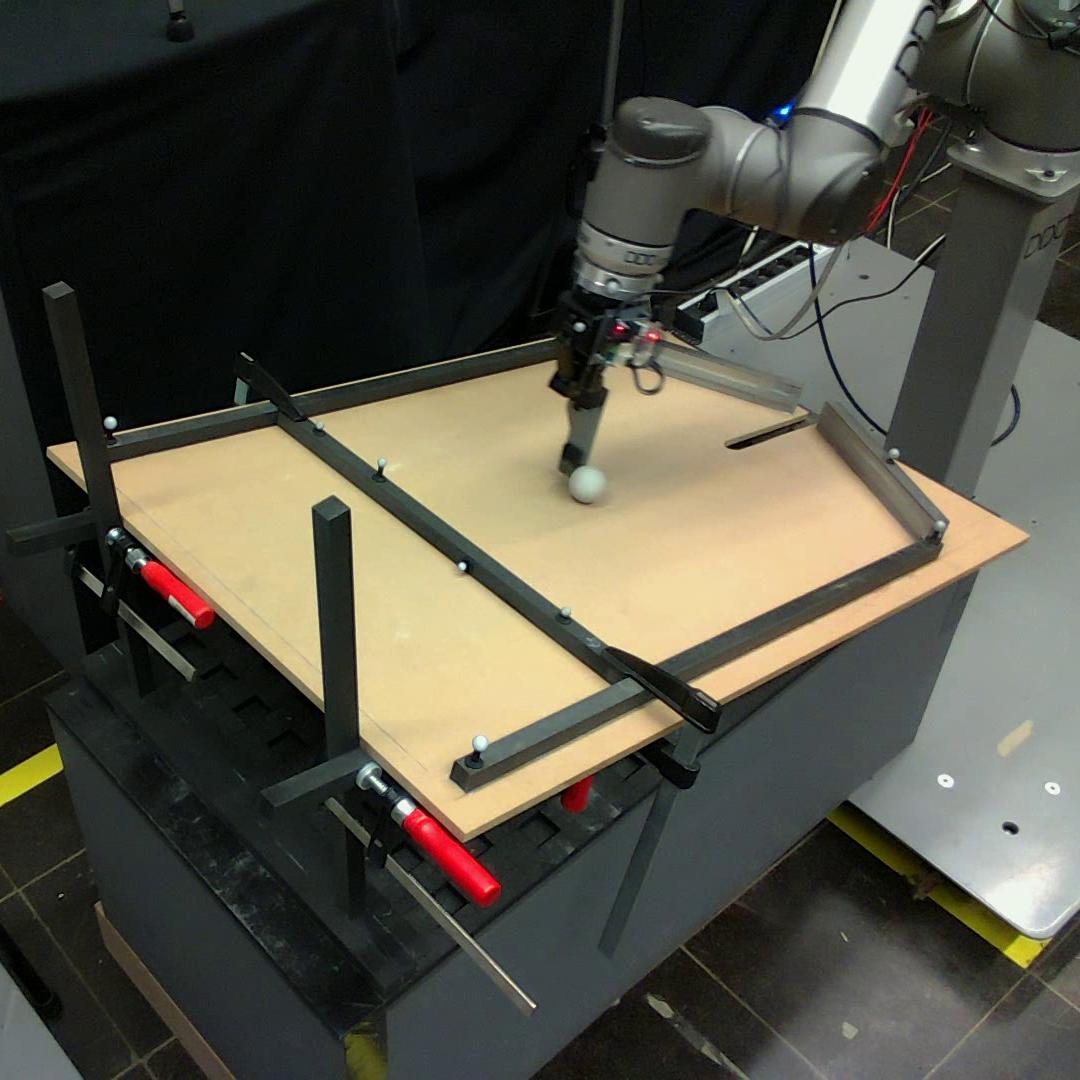} &
        \includegraphics[width=\linewidth]{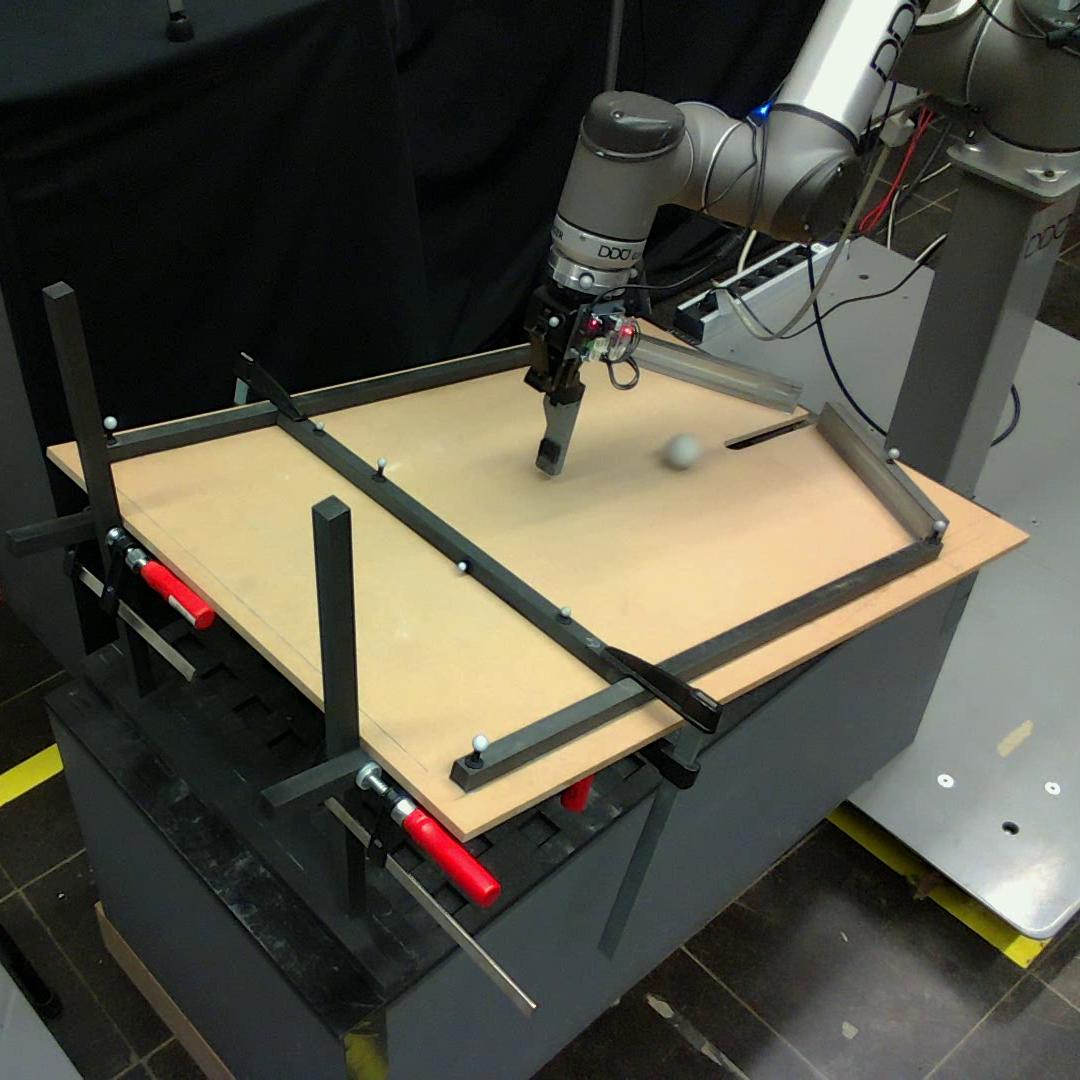} &
        \includegraphics[width=\linewidth]{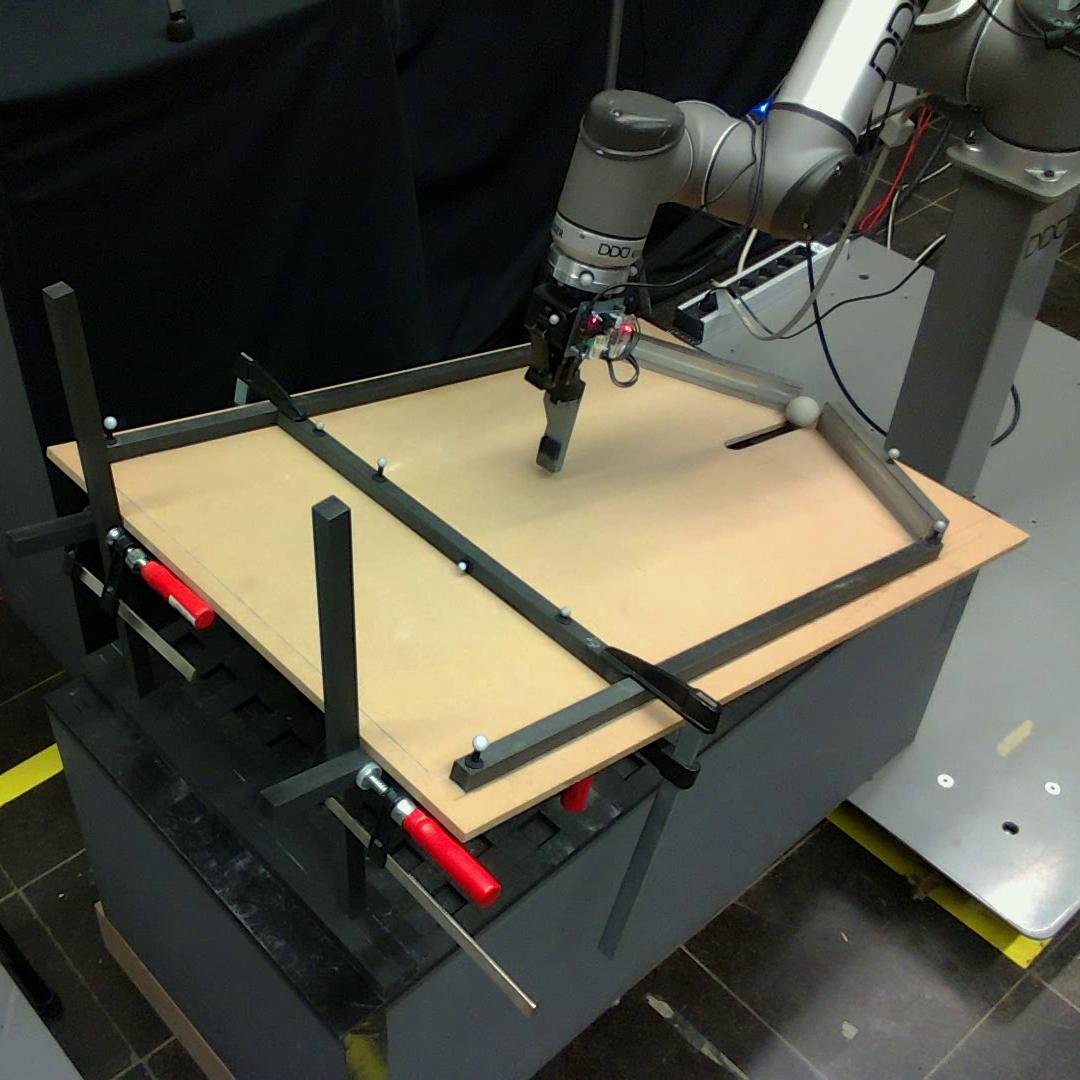}
        \\[1.10cm]
        30k steps &
        \includegraphics[width=\linewidth]{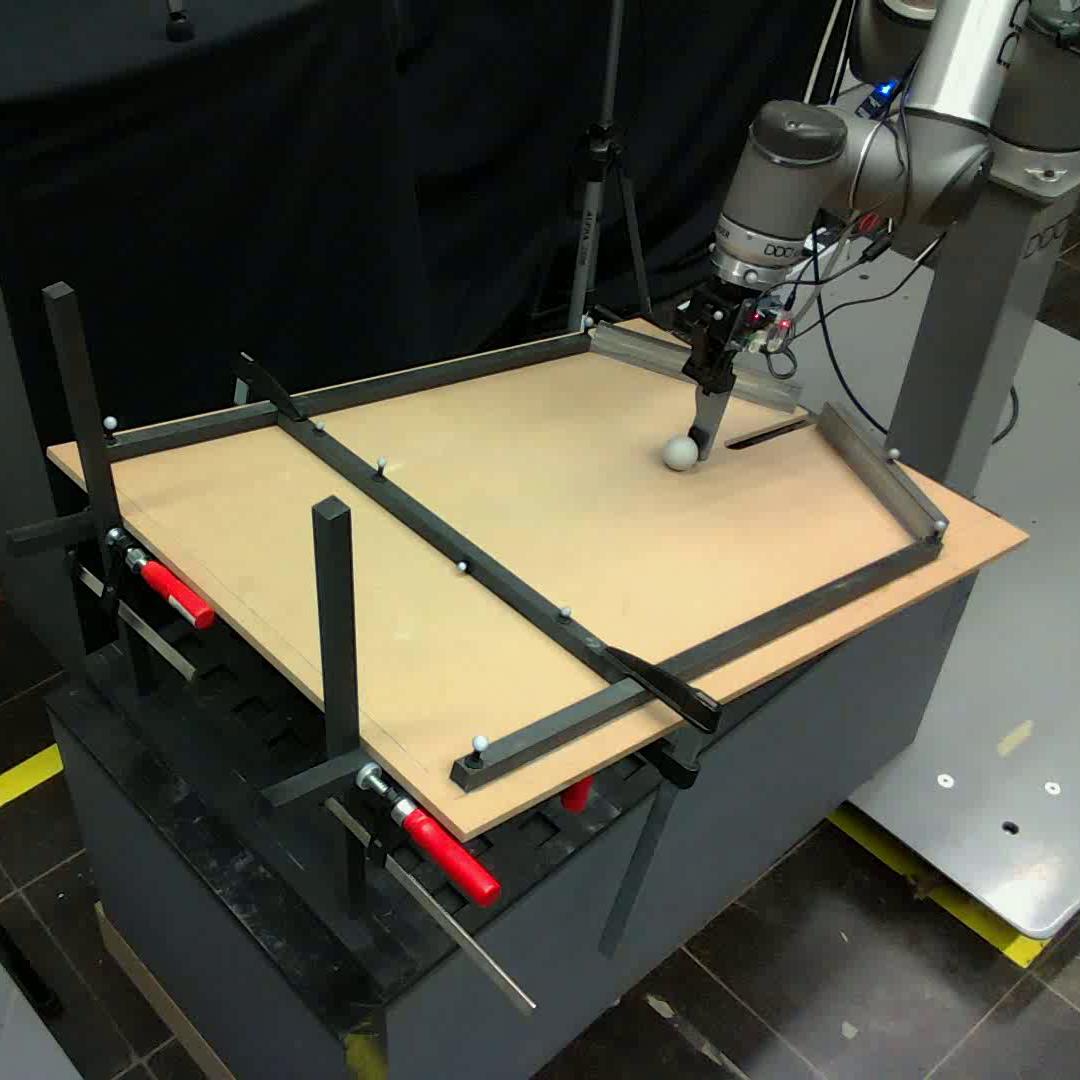} &
        \includegraphics[width=\linewidth]{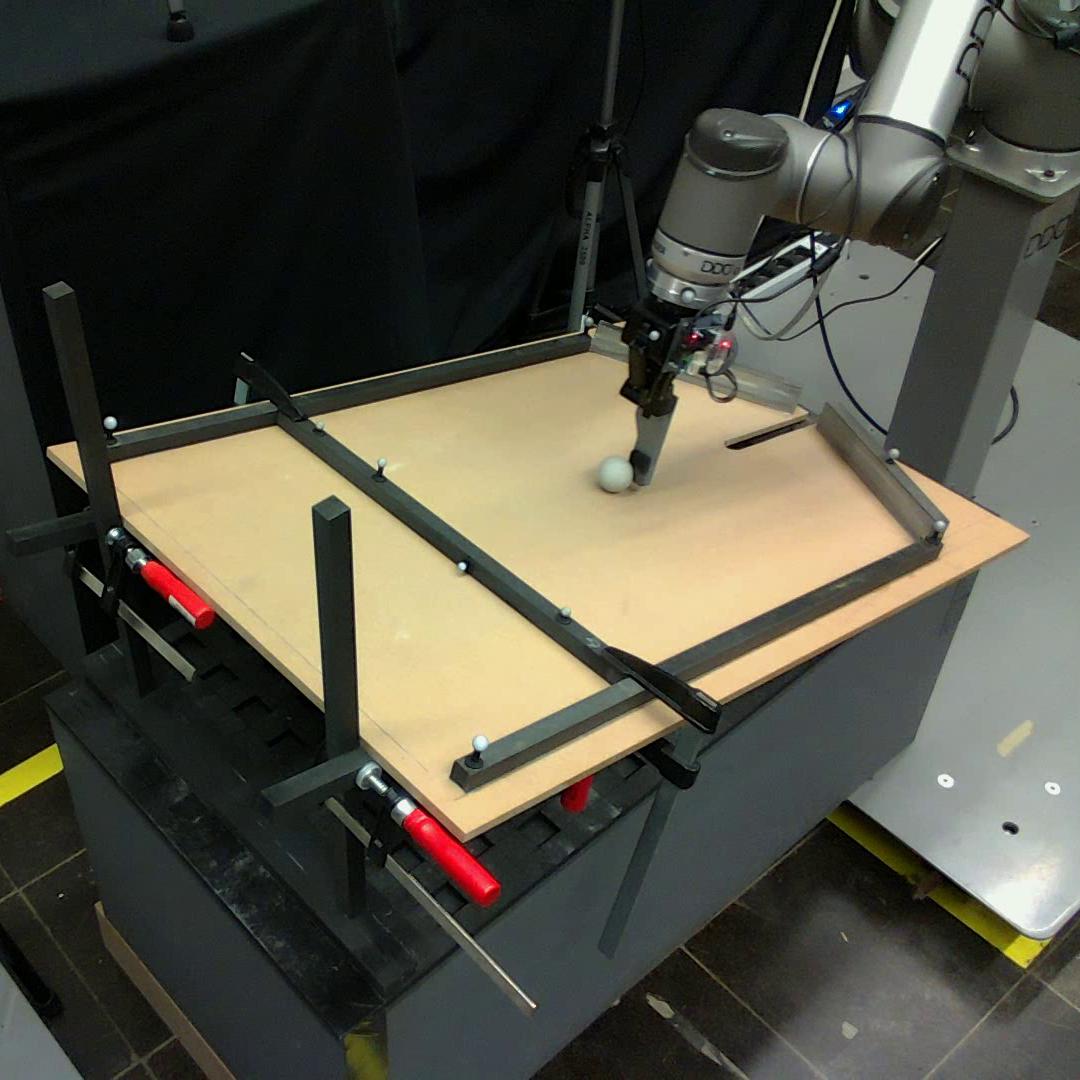} &
        \includegraphics[width=\linewidth]{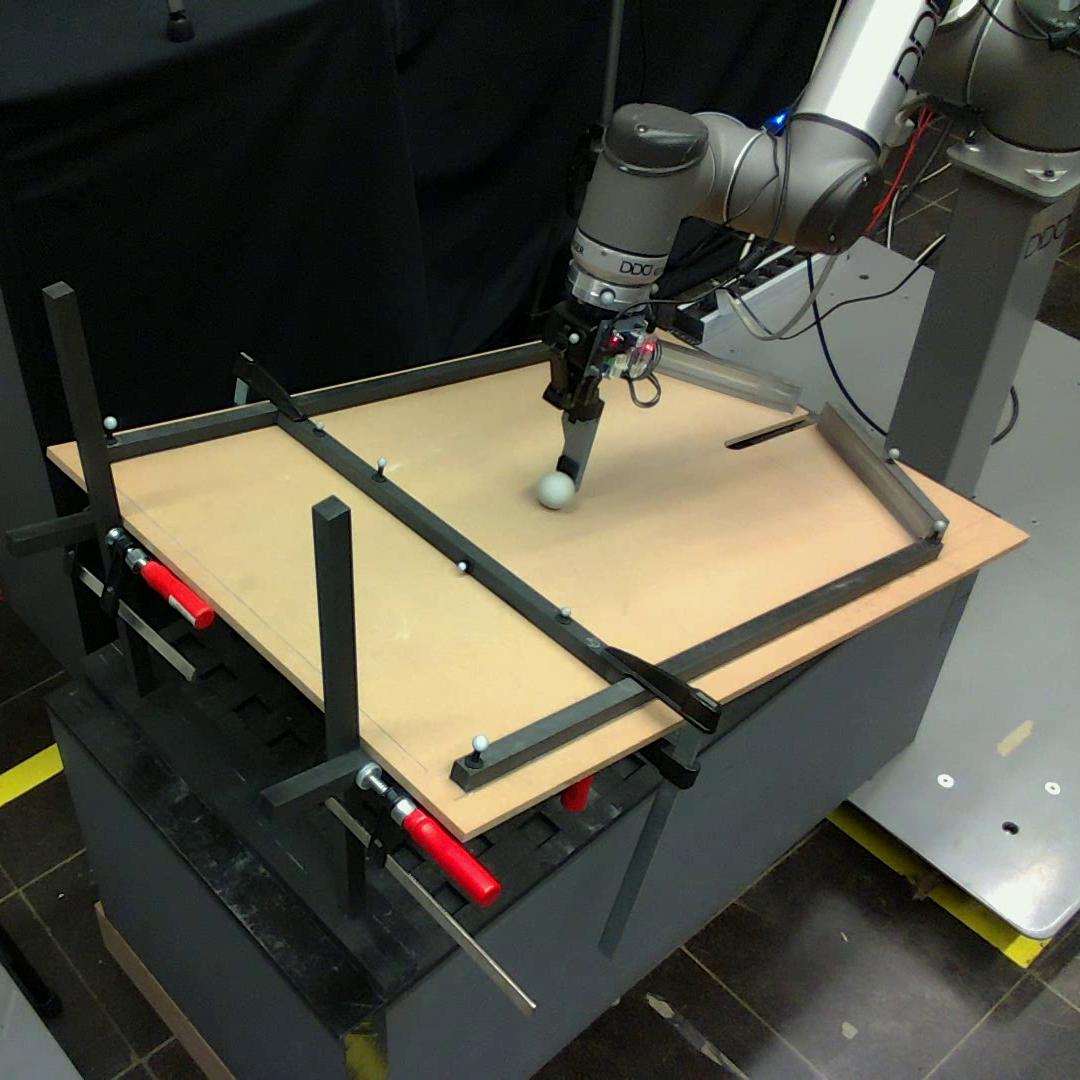} &
        \includegraphics[width=\linewidth]{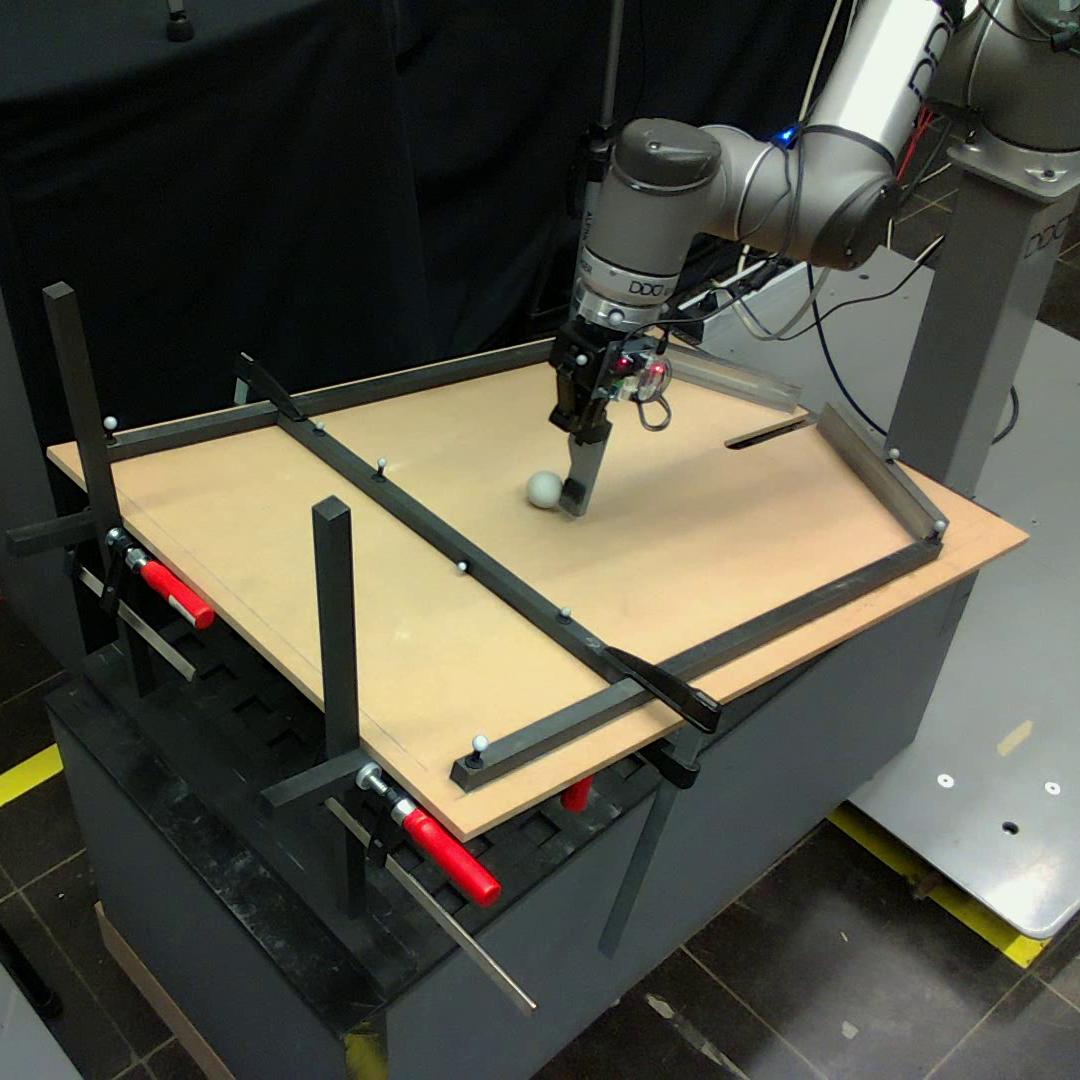} &
        \includegraphics[width=\linewidth]{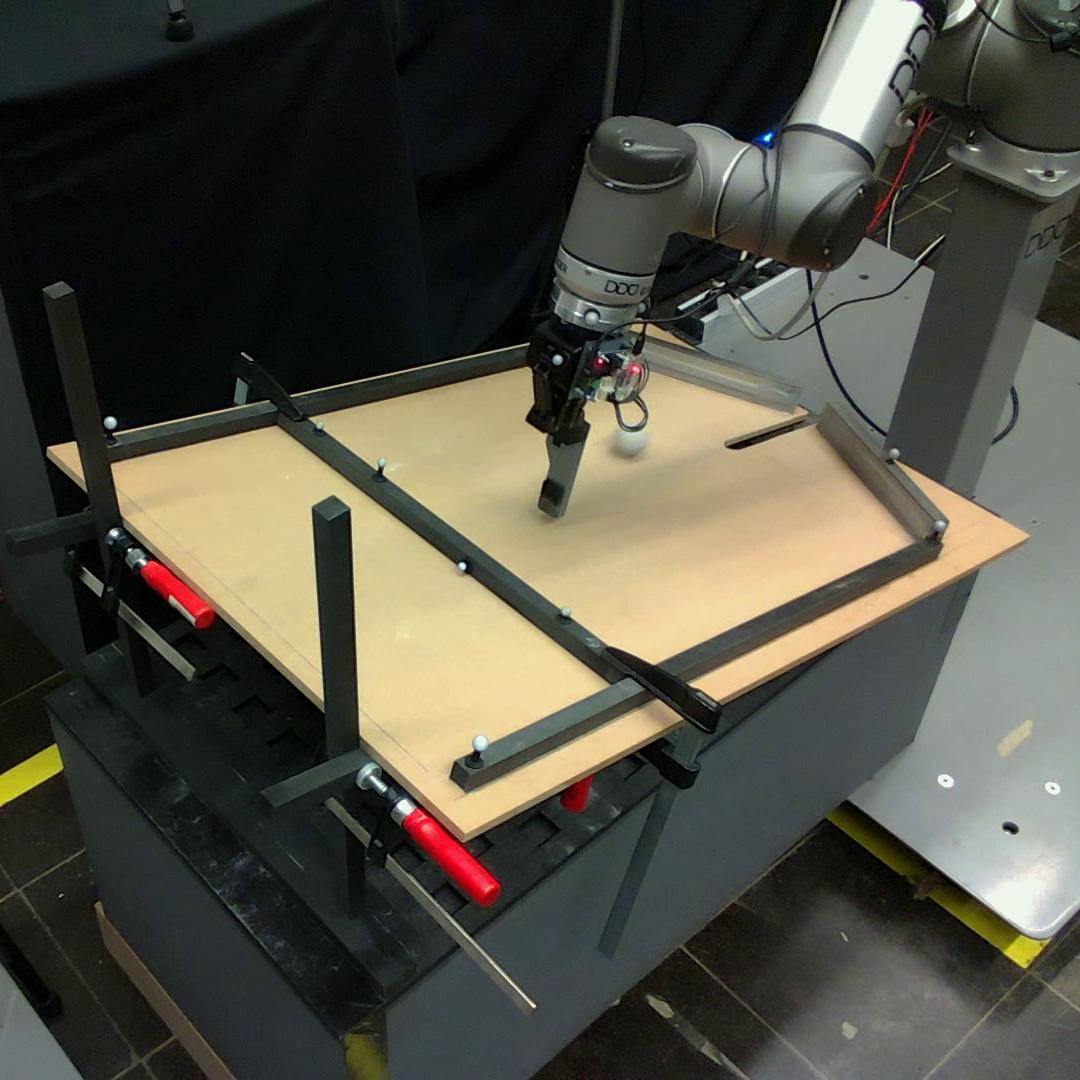}
        \\[1.10cm]
        150k steps &
        \includegraphics[width=\linewidth]{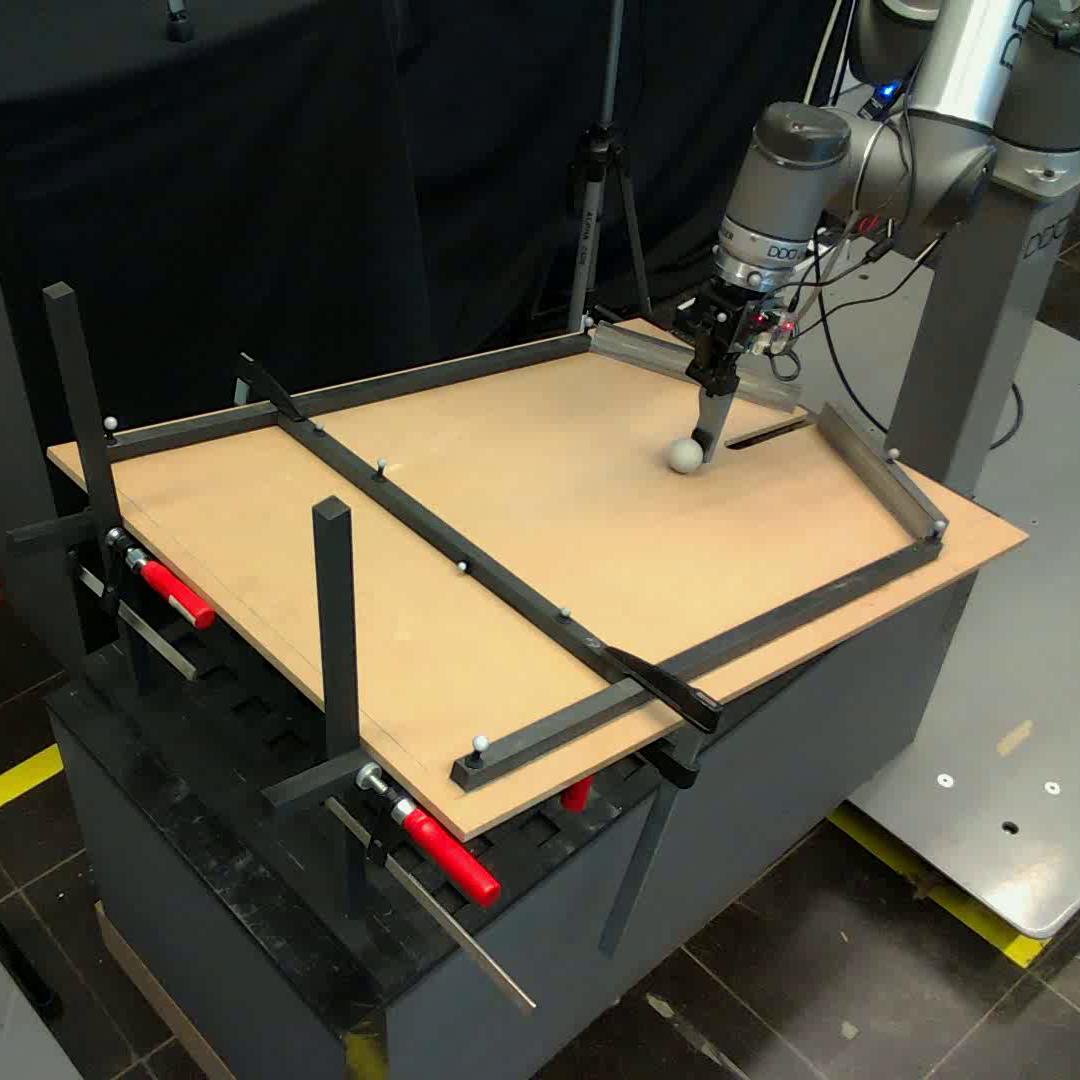} &
        \includegraphics[width=\linewidth]{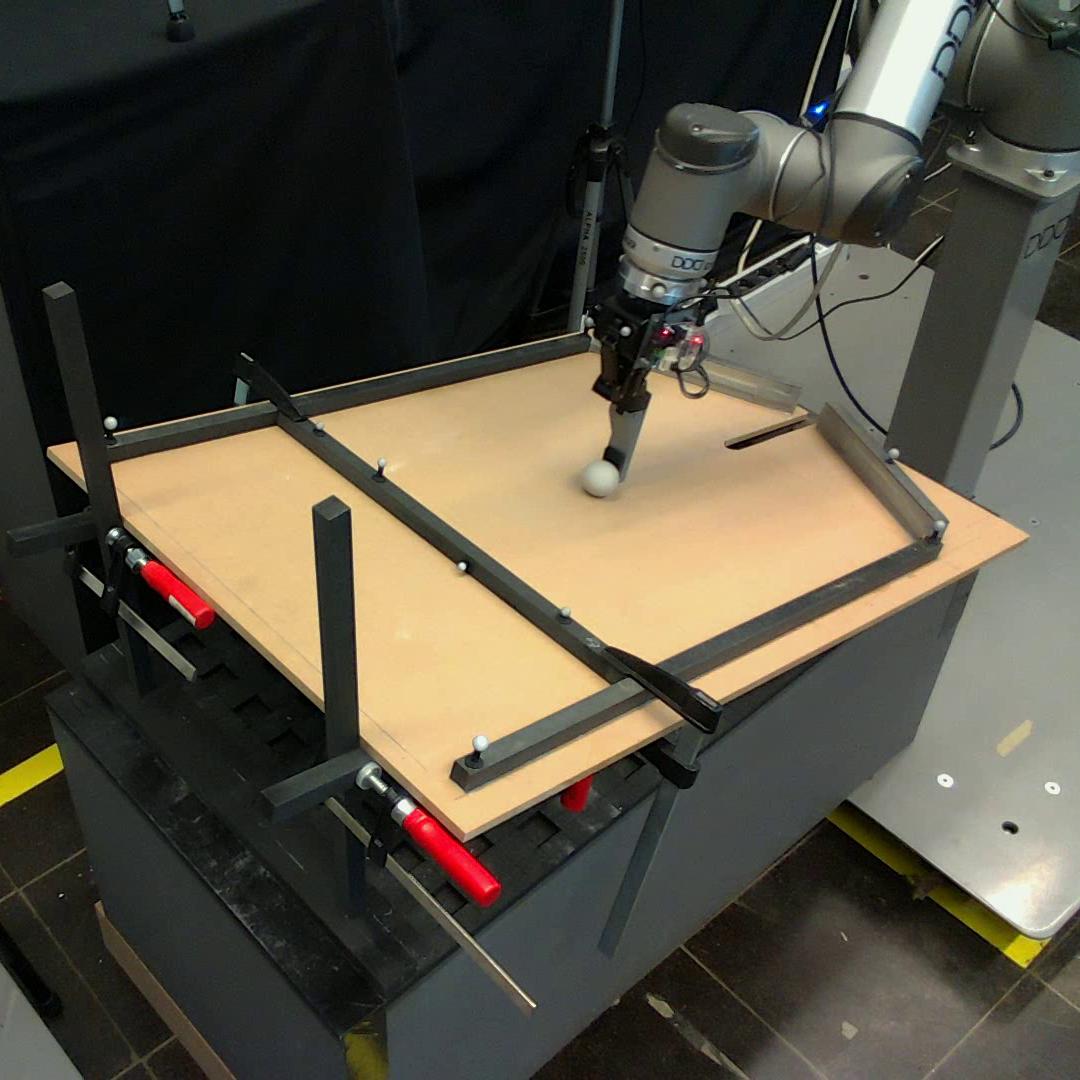} &
        \includegraphics[width=\linewidth]{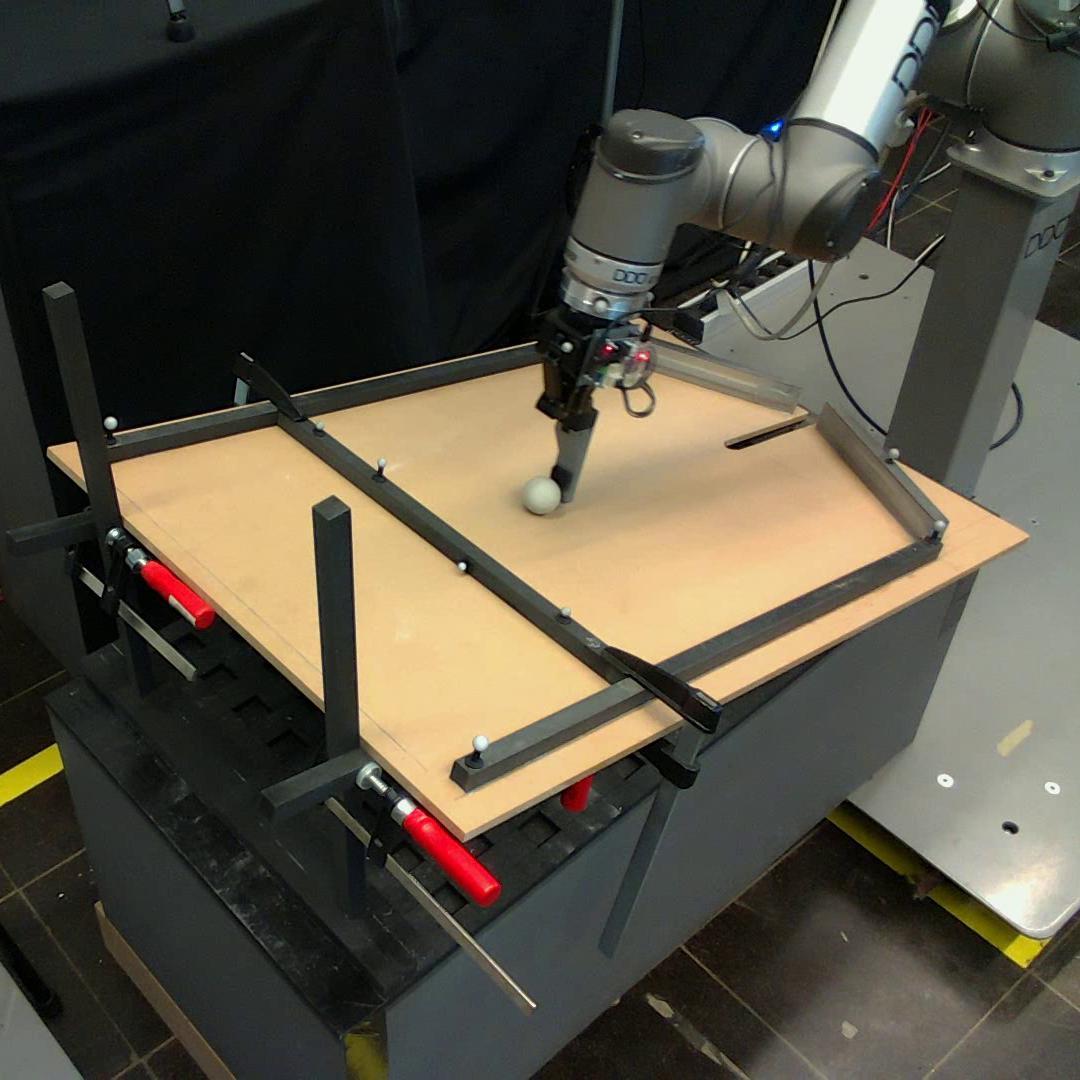} &
        \includegraphics[width=\linewidth]{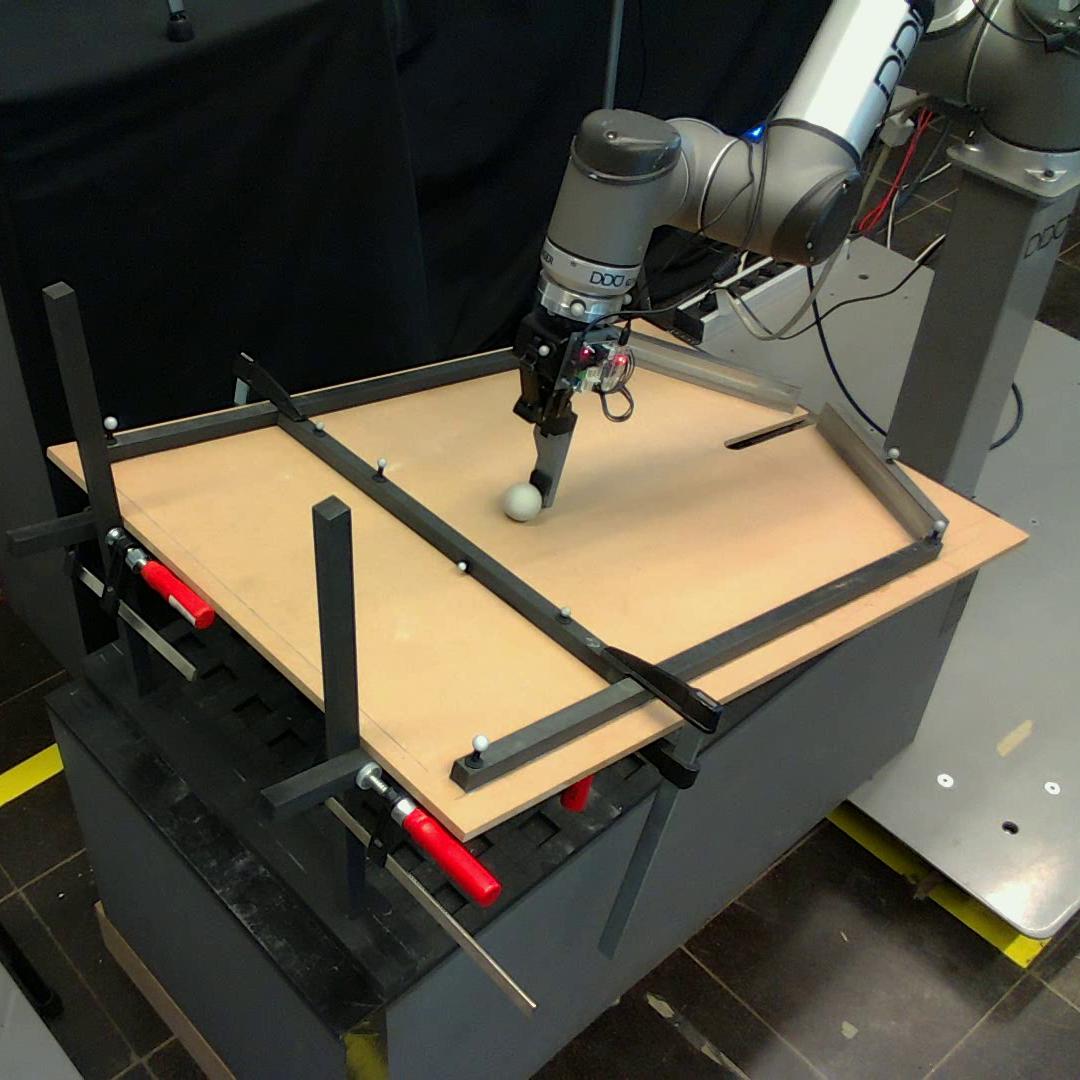} &
        \includegraphics[width=\linewidth]{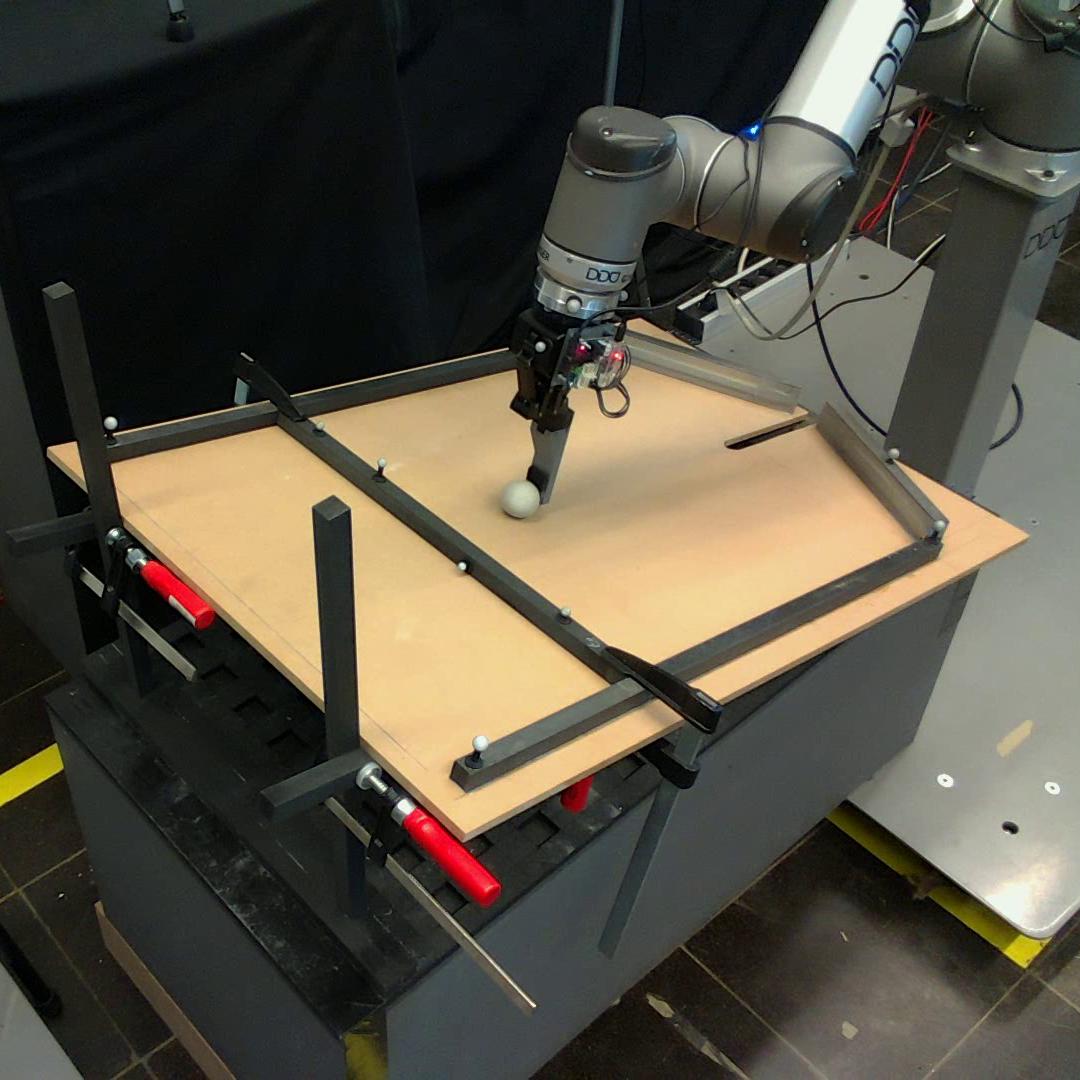}
    \end{tabular}\setlength\tabcolsep{6pt}
    \caption{
        Training process of our agent on the real system.
        At 5k steps, the agent has not found the target yet, and thus it is not guided by any external reward signal.
        Instead, purely driven by its intrinsic drive, it learns to balance the ball on the finger and systematically explores the table.
        At 30k steps, the agent understood how to obtain the reward in this task and the extrinsic signal causes it to focus on exploitation from now on.
        At 150k steps, the agent's model is accurate enough to repeatedly balance the ball at the goal location and solve the task.
    }
    \label{fig:real_vis}
    \vspace{-1em}
\end{figure*}

\cref{fig:exp_ball_res_real} shows the results of training with our method on the real-world setup.
Note that we do not pre-train in simulation, but learn the model from scratch in the real world.
Despite the state not being fully observable, our method solves the \textit{\balltaskname} in the real-world with a performance comparable to the simulated equivalent.

These experiments show that our method is able to systematically explore a complex, contact-rich environment with many dead-ends.
Without any extrinsic feedback, our agents learned to balance the ball on the end-effector and to systematically move it around the environment until the target zone is found.
The sole reason for this behavior to occur in the first place is that our agents understood they could only explore the entire state-space, if they kept balancing the ball and move it to unseen locations.
Our final experiment shows that our method generalizes to the real world, while only requiring some simple algorithmic changes.
These results serve as a proof of concept for using model-based active exploration to learn challenging robotic manipulation tasks.

%% file: s4_active_inference.tex
\section{Connections to \texorpdfstring{\glsxtrlong{ai}}{Active Inference}}

Our approach is related to the \glsfirst{ai}~\cite{parr2022active} framework, which we briefly summarize in the following.
\gls{ai} is an implementation of the \gls{fep}~\cite{friston2010action}, which attempts to explain intelligent behavior from a cognitive science perspective.
The fundamental idea behind \gls{fep} is that any organism's objective is to restrict the states it is visiting to a manageable amount.
From this basic principle, a process theory is derived that reproduces features of intelligent behavior and curiosity~\cite{friston2015active}.

Mathematically, \gls{ai} implements the \gls{fep}'s objective as follows:
An agent maintains a generative model $p$ of the world and avoids sensations $o$ that are surprising, i.e., that have a low marginal log-probability $\ln \p{\vo}$.
Thus, the objective can be written as
\begin{align}
    \min_{\pi}\; -\ln \p{\vo}
\end{align}
where $o$ is generated by an external process that can be influenced by changing the policy $\pi$.

The agent's generative model is assumed to consist not only of observations $o$, but also include hidden states $\vh$, giving $\p{\vo} = \int \p{\vo, \vh} d\vh = \int \p{\vo}[\vh] \p{\vh} d\vh$.
To make optimization tractable, variational inference is invoked to obtain the \glsxtrlong{elbo} using Jensen's inequality:
\begin{align}
    -\ln \p{\vo}
    &=
    -\ln \int \p{\vo, \vh} d\vh
    =
    -\ln \int \frac{\q[\phi]{\vh}}{\q[\phi]{\vh}} \p{\vo, \vh} d\vh
    \\
    &\leq
    \kl{\q[\phi]{\vh}}{\p{\vh}[\vo]} - \ln \p{\vo}
    \eqqcolon
    \vfesym \left( \vo, \phi \right)
\end{align}
where $\vfesym \left( \vo, \phi \right)$ is termed the \gls{vfe}.

Minimizing $\vfesym \left( \vo, \phi \right)$ w.r.t.\ the variational parameters $\phi$ corresponds to minimizing the \glsxtrshort{kl} divergence between the variational posterior $\q[\phi]{\vh}$ and the true posterior $\p{\vh}[\vo]$.
In other words, by minimizing the \gls{vfe} w.r.t.\ $\phi$, the agent is solving the perception problem of mapping its observations to their latent causes.

To facilitate planning into the future, the \gls{vfe} can be modified to incorporate an expectation over future states, yielding the \gls{efe}~\cite{friston2015active}:
\begin{align}
    \efesymP_{\pi} \left( \phi \right)
    &=
    - \Ex{\q[\phi]{\mathbf{\vo}, \mathbf{\vh}}[\pi]}{
        \ln \p{\mathbf{\vo}, \mathbf{\vh}}
        - \ln \q[\phi]{\mathbf{\vh}}[\pi]
    }
    \\
    &\approx
    -
    \sum_{\tau = t + 1}^{t+H}
    \Ex{\q[\phi]{\vo_\tau, \vh_\tau}[\pi]}{
        \ln \p{\vo_\tau, \vh_\tau}
        - \ln \q[\phi]{\vh_\tau}[\pi]
    }
    \\
    \label{eq:efe}
    \begin{split}
        &\approx
        -
        \sum_{\tau = t + 1}^{t+H}
        \Big(
        \underbrace{
            \Ex{\q[\phi]{\vo_\tau}[\pi]}{\ln \p{\vo_\tau}}
        }_{\text{extrinsic term}}\rhookswarrow
        \\
        &\quad + \underbrace{
            \Ex{\q[\phi]{\vh_\tau}[\pi]}{
                \kl{\q[\phi]{\vo_\tau}[\vh_\tau, \pi]}{\q[\phi]{\vo_\tau}[\pi]}
            }
        }_{\text{intrinsic term (expected information gain)}}
        \Big)
    \end{split}
\end{align}
where a mean-field assumption is made in the second step and definitions $\mathbf{\vo} \coloneqq o_{t+1:t+H}$ and $\mathbf{\vh} \coloneqq \vh_{t+1:t+H}$ are used.

The minimization of the \gls{efe} w.r.t.\ the policy $\pi$ causes the agent to act in a way that maximizes both the information gain and the extrinsic term.
Here, the extrinsic term acts as an external signal expressing the preferences of the agent over observations.
While it is common in \gls{rl} to use a reward function to give the agent a notion of ``good'' and ``bad'' behavior, in the \gls{ai} framework, one defines a prior distribution over the target observations $\p{\vo}$ that the agent tries to match.
By making the reward part of the observation and setting the maximum reward as the target observation, one can transform any reward-based task to fit into the \gls{ai} framework~\cite{tschantz2020reinforcement}.

In our implementation, the agent observes the state of the environment~$\vs_\tau$ and the reward~$\vr_\tau$ at every time step~$\tau$.
The only unobserved variables are the model parameters~$\theta$.
Consequently, the hidden state is given by $\vh_\tau = \left( \vs_\tau, \vr_\tau \right)$.
Setting the preference distribution to $\p{o_\tau} \propto \exp(\beta r_\tau)$ and dropping constant terms makes the \gls{efe} at time $t$ become
\begin{align}
    \label{eq:efe_real}
    \begin{split}
        &\efesymP_{\pi} \left( \phi \right)
        \propto
        -
        \sum_{\tau = t + 1}^{t + H}
        \Big(
        \beta \Ex{\q[\phi]{\vr_\tau}[\pi]}{\vr_\tau}\rhookswarrow
        \\
        &\quad +
        \Ex{\q[\phi]{\theta}[\pi]}{
            \kl{\q[\phi]{\vh_\tau, \vr_\tau}[\theta, \pi]}{\q[\phi]{\vh_\tau, \vr_\tau}[\pi]}
        }
        \Big).
    \end{split}
\end{align}
The only difference between the above objective and our planning objective is that we do not make the mean-field assumption over time.
Hence, our method can be understood as an implementation of \gls{ai} where the agent encodes its belief about the dynamics of the environment in its hidden state.
As such our method is related to \cite{tschantz2020reinforcement}, where a variant of \gls{ai} was used for model learning and the results were reported for simple benchmark environments, such as \textit{Mountain Car}~\cite{brockman2016openai} and \textit{Cup Catch}~\cite{tassa2020dmcontrol}.
To our knowledge, our method is the first demonstration of \gls{ai} used for active model learning on a real robotic system on a sparse-reward manipulation task.

%% file: s5_conclusion.tex
\section{Conclusion}

In this paper, we developed an active exploration method that is capable of solving complex robotic manipulation tasks.
Our main algorithmic contribution is the introduction of an information-seeking strategy in model-based reinforcement learning, that balances between exploration of new states in the environment to improve the dynamics model and task performance.
We evaluated our method on two simulated and one real-world tasks, all designed to be particularly hard to explore.
Throughout our experiments, we showed that our method induces systematic exploratory behavior and learns to solve a manipulation task without dense extrinsic reward, but is driven by its own curiosity.
Considering that none of the baselines were able to solve these problems, we conclude that the information-seeking behavior of our agents is beneficial for solving challenging exploration problems with sparse rewards, suitable for learning complex manipulation tasks in the real world.

In the future, we plan to incorporate tactile sensors into our setup.
Tactile sensors would allow one to obtain more detailed feedback about the objects being manipulated.
For example, the spin of the ball was not observed in our experiments, although it provides a useful signal for the manipulation task.
Considering that humans deploy a variety of active haptic exploration strategies during manipulation, research on robotic active tactile exploration might bring us closer to human-level manipulation skills.
One of the main limitations of our method that prevents it from being used on more dynamic tasks is that it is computationally comparably heavy and therefore limited to relatively slow tasks.
Hence, an exciting future research direction is to tackle this issue with hierarchical controllers by combining our planning module with a learned low-level balancing controller running at a high frequency.

%% file: s6_appendix.tex
\clearpage
\onecolumn
\setlength{\parindent}{0pt}
\setlength{\parskip}{0.5em}
\appendix

\subsection{Full algorithm}
On a high level, our algorithm alternates between collecting new roll-outs and fitting the ensemble models to the data that has been observed during these roll-outs.
During roll-outs, we rely on \gls{mpc} to choose continuous actions.
That is, in every step we use our planner to generate a plan for a horizon of 20 time steps and execute the first action of this plan on the system.
More details on the planner can be found in \cref{subsec:app_planner}.
For details on how we fit the ensemble models to the observed data, refer to \cref{subsec:app_model}.
The full pseudo-code of our algorithm is given in \cref{alg:app_full_algo}.

\begin{algorithm}[H]
    \begin{algorithmic}[1]
        \State Initialize replay buffer
        \State Initialize ensemble parameters $\theta^1, \dots, \theta^P$ randomly
        \For{$e = 1, \dots, N$}
            \State Reset agent and obtain initial state $x_0$
            \For{$t = 1, \dots, T_{\max}$}
                \State $\va_t \leftarrow \argmax_a \Ex{\p{\mathbf{\vr}_{t+1:T}}[\pi]}{\sum_{\tau = t + 1}^T \vr_\tau} + \beta \text{I} \left( \pi, \vs_t \right)$
                \State Execute $\va_t$ and obtain new state $\vs_t$ and reward $\vr_t$
                \State Store $\left( \vs_t, \vr_t, \va_t \right)$ in replay buffer
            \EndFor
            \State Optimize $\theta^1, \dots, \theta^P$ using the replay buffer
        \EndFor
    \end{algorithmic}
    \caption{Goal-Directed Active Exploration via MPC}
    \label{alg:app_full_algo}
\end{algorithm}

\subsection{Planning algorithm}
\label{subsec:app_planner}

As noted before, we use a variant of the \glsfirst{cem} to find action sequences $\mathbf{\va}_{t+1:T}$ that maximize the planning objective \cref{eq:obj} in real time.
\gls{cem} is a sampling-based optimization algorithm that solves problems of the form
\begin{align}
    \min_{\mu, \sigma} \quad
    \Ex{\mathcal{N} \left( \nu \mid \mu, \text{diag} \left( \sigma \right) \right)}{f \left( \nu \right)}
\end{align}
where $f \colon \R^{N_\nu} \rightarrow \R$ is an objective function depending on some variable $\nu \in \R^{N_\nu}$ and $\mu, \sigma \in \R^{N_\nu}$ are the parameters of Gaussian distribution with diagonal covariance matrix.
Hence, \gls{cem}'s objective is to find the parameters of a Gaussian distribution over $\nu$, that maximize the expectation of the objective function $f \left( \nu \right)$.
In our case, $\nu$ is the sequence of future actions $\mathbf{\va}_{t+1:T}$, concatenated into a single vector.

The way this problem is approached by \gls{cem} is shown in \cref{alg:cem}.
It can be summarized as follows:
First, $\mu$ and $\sigma$ are initialized to $\mu = 0$ and $\sigma = \mathbb{1}$.
Then, at every iteration, $N$ samples $\nu_1, \dots, \nu_N$ are drawn from the current distribution.
Out of these samples, the $M$ samples with the highest function value under $f$ are selected and $\mu$ and $\sigma$ are fitted to these samples.
This way, the current distribution is gradually shifted towards areas of $f$ that hold a high value.

\begin{algorithm}[H]
    \begin{algorithmic}[1]
        \State Initialize $\mu \leftarrow 0, \sigma \leftarrow \mathbbm{1}$
        \For{$i = 1, \dots, n$}
            \State Sample $\mathbf{\va}_1, \dots, \mathbf{\va}_N \sim \mathcal{N} \left( \mu, \text{diag} \left( \sigma \right) \right)$
            \State Evaluate reward $f_{j} \leftarrow f \left(\mathbf{\va}_j \right) \quad \forall j \in \left\{ 1, \dots, N \right\}$
            \State Collect the $M$ samples with highest $f_j$: $\mathbf{\va}^\ast_1, \dots, \mathbf{\va}^\ast_M$
            \State Compute new means and standard deviations:

            $\mu \leftarrow \frac{1}{M} \sum_{j = 1}^{M} \mathbf{\va}^\ast_j$

            $\sigma \leftarrow \sqrt{\frac{1}{M} \sum_{j = 1}^{M} \left( \mu - \mathbf{\va}^\ast_j \right)^T \left( \mu - \mathbf{\va}^\ast_j \right)}$
        \EndFor\\
        \Return $\mu_1$, $\sigma_1$
    \end{algorithmic}
    \caption{Vanilla Cross Entropy Method for planning}
    \label{alg:cem}
\end{algorithm}

\subsubsection{Overcoming \textit{detachment} with a memory buffer}
While standard \gls{cem} has been shown to achieve great results in a variety of tasks~\cite{hafner2019learning}, we found it to be unfit to deal with the extremely sparse-reward tasks we presented in \cref{sec:exp}.
The issue we encountered is that although vanilla \gls{cem} does well in the beginning of the training, exploring the state space around the initial state thoroughly, it stops exploring at some point and converges to the locally optimal strategy of not moving to avoid the action penalty.
This behavior is caused by an effect that has been termed \textit{detachment} in prior work~\cite{Ecoffet2019Jan}~\footnote{%
    Note that the definition of detachment \cite{Ecoffet2019Jan} give is not in the context of purely model-based reinforcement learning and, thus, differs slightly from the issue we encounter here.
    However, we believe that our issue is the model-based equivalent of the definition they give.
}.

In our case, \textit{detachment} occurs when the agent has thoroughly explored the state space around the initial state distribution, causing the intrinsic reward of this area to become low.
Once the intrinsic reward is close to zero around the initial state space, it becomes increasingly difficult for the \gls{cem} planner to find a trajectory that leads to the frontier of unexplored states, where the intrinsic reward is high again.
The reason for this increase in difficulty is that in the absence of both extrinsic and intrinsic reward around the current state of the agent, there is no reward signal guiding the planner out of the already explored area.
Hence, unless we are lucky and sample a trajectory that leads into unexplored regions of the state space, all trajectories generated during planning will have a similar reward, giving the planner no indication in which direction it should shift the trajectory distribution.
In other words, as soon as the intrinsic reward around the initial state distribution diminishes, the planner has to find its way out of a plateau in the objective function, which is often not possible for local optimization methods like \gls{cem}.
Especially in the case of our environments, escaping these plateaus can be particularly hard, as it usually requires to balance the ball somewhere, which is only possible with a very narrow set of action sequences.
During our experiments we found that finding those sequences by pure chance is very unlikely, rendering our method with the vanilla \gls{cem} planner unable to solve the tasks we presented in \cref{sec:exp}.

A way we found to be very effective in resolving this issue is to use trajectories planned in the past to warm start the planner.
The reason this modification is so effective is that in the set of past trajectories there must be trajectories leading to the frontier of unexplored states, because at some point the frontier was pushed to the place it is currently.
Hence, instead of only sampling trajectories randomly in the first iteration of \gls{cem}, we also sample trajectories planned in the past.
Technically, we realize this modification to \gls{cem} by implementing a memory buffer, in which we store the current state and the parameters of the planned action trajectory distribution at the end of each \gls{cem} execution.
Then, in the first iteration of the following \gls{cem} executions, we search the K nearest neighbors in terms of the current state $s_t$ in this buffer and use the resulting action trajectory distributions to sample additional initial trajectories from.
The modified \gls{cem} algorithm is given in \cref{alg:cem_mod}, with the modifications marked in red.

Both the detachment problem and our proposed solution are visualized in \cref{fig:detachment}.

\begin{algorithm}[H]
    \begin{algorithmic}[1]
        \State Initialize $\mu \leftarrow 0, \sigma \leftarrow \mathbbm{1}$
        \For{$i = 1, \dots, n$}
            \State Sample $\mathbf{\va}_1, \dots, \mathbf{\va}_N \sim \mathcal{N} \left( \mu, \text{diag} \left( \sigma \right) \right)$
            \If{i = 1}
                \State \textcolor{red}{Fetch K-NN of $s_t$ from memory buffer $\mathcal{B}$: $(\hat{\mu}_1, \hat{\sigma}_1), \dots, (\hat{\mu}_K, \hat{\sigma}_K)$}
                \State \textcolor{red}{Additionally sample
                    $\mathbf{\va}_{N + k S + j} \sim \mathcal{N} \left( \hat{\mu}_k, \text{diag} \left( \hat{\sigma}_k \right) \right)
                    \forall k \in \left\{ 1, \dots, K \right\} \forall j \in \left\{ 1, \dots, S \right\}$}
            \EndIf
            \State Evaluate reward $f_{j} \leftarrow f \left(\mathbf{\va}_j \right) \quad \forall j \in \left\{ 1, \dots, N (+ KS) \right\}$
            \State Collect the $M$ samples with highest $f_j$: $\mathbf{\va}^\ast_1, \dots, \mathbf{\va}^\ast_M$
            \State Compute new means and standard deviations:

            $\mu \leftarrow \frac{1}{M} \sum_{j = 1}^{M} \mathbf{\va}^\ast_j$

            $\sigma \leftarrow \sqrt{\frac{1}{M} \sum_{j = 1}^{M} \left( \mu - \mathbf{\va}^\ast_j \right)^T \left( \mu - \mathbf{\va}^\ast_j \right)}$
        \EndFor
        \State \textcolor{red}{Store $(s_t, \mu, \sigma)$ in memory buffer $\mathcal{B}$}\\
        \Return $\mu_1$, $\sigma_1$
    \end{algorithmic}
    \caption{Cross Entropy Method with memory buffer}
    \label{alg:cem_mod}
\end{algorithm}

\subsubsection{Hyperparameter choices for the planner}

We run our modified \gls{cem} for $n \coloneqq 12$ iterations in every experiment.
The number of samples per iteration $N$ we set to 500, the number of fetched neighbors $K$ to 50, the samples per neighbor $S$ to 10, and the number of elites $M$ to 20.
That means in the first iteration, the population of action trajectories comprises 500 random samples and 500 samples sampled from past plans.
Furthermore, we set the size of the memory buffer $\mathcal{B}$ to 50,000 action trajectories.
To ensure that we can search through this buffer quickly, we utilize the efficient KNN implementation of \texttt{faiss}~\cite{johnson2019billion}.

\definecolor{detFog}{HTML}{7f7f7f}
\definecolor{detDisc}{HTML}{F8BA3C}
\definecolor{detDiscCur}{HTML}{F5A300}
\definecolor{detNewPlan}{HTML}{A40A0A}

\newcommand{\drawCircles}[2][detDiscCur]{
    \foreach \p [count=\i,remember=\p as \lastp] in #2{
        \ifnum\i>1\relax
        \draw[line width=1.0cm,line cap=round,#1] \lastp -- \p;
        \fi
    }
}

\NewDocumentCommand\drawPath{O{1.0}mO{black}}{%
    \foreach \p [count=\i,remember=\p as \lastp] in #2{
        \ifnum\i>1\relax
        \draw[-latex,line width=0.3mm,opacity=#1,color=#3] \lastp -- \p;
        \fi
    }
}

\newcommand{\ifthen}[2]{
    \ifnum #1
    #2
    \fi
}

\def\expPathA{(-3, 0), (-2.5, 0.5), (-2, 0), (-1.7, 0.7), (-2.5, 1.2), (-2.4, 2.2), (-3.7, 2.1), (-4.5, 2.5)}
\def\expPathB{(-3, 0), (-3.1, 0.7), (-3.2, 1.5), (-4.2, 1.2), (-3.9, -0.1), (-4.4, -0.8), (-4.2, -1.8), (-4.3, -2.6)}
\def\expPathC{(-3, 0), (-3.5, -0.9), (-3.4, -2.1), (-2.7, -2.3), (-1.5, -2.1), (-1.2, -1.0), (-0.9, -0.1), (0.6, 0.1)}
\def\expPathD{(-3, 0), (-2.7, -1.3), (-2.1, -1.4), (-1.9, -0.3), (-1.0, 0.3), (-1.5, 1.2), (-1.7, 2.2), (-1.0, 2.0)}
\def\expPathL{(-3, 0), (-2.3, 0.1), (-1.7, 0.0), (-1.5, 0.6), (-2.0, 1.0), (-2.7, 0.7), (-3.0, 1.1), (-2.6, 1.8)}
\def\expPathE{(-3, 0), (-1.9, -0.3), (-1.3, 0.4), (-0.6, 1.3), (1.0, 1.4), (2.0, 0.3), (2.4, 1.4), (3.4, 0.2)}
\newcommand{\drawDet}[1]{
    \begin{tikzpicture}
        \begin{scope}
            \node[rectangle,fill=detFog,minimum width=9.5cm, minimum height = 5.5cm] (bg) at (0, 0) {};
            \begin{pgfinterruptboundingbox}
                \begin{scope}
                    \clip (-4.75, -2.75) rectangle (4.75, 2.75);
                    \ifthen{#1=1}{\drawCircles{\expPathA}}
                    \ifthen{#1>1}{\drawCircles[detDisc]{\expPathA}}
                    \ifthen{#1=2}{\drawCircles{\expPathB}}
                    \ifthen{#1>2}{\drawCircles[detDisc]{\expPathB}}
                    \ifthen{#1=3}{\drawCircles{\expPathC}}
                    \ifthen{#1>3}{\drawCircles[detDisc]{\expPathC}}
                    \ifthen{#1=4}{\drawCircles{\expPathD}}
                    \ifthen{#1>4}{\drawCircles[detDisc]{\expPathD}}
                    \ifthen{#1=7}{\drawCircles[detDisc]{\expPathE}}
                    \ifthen{#1>7}{\drawCircles{\expPathE}}
                \end{scope}
            \end{pgfinterruptboundingbox}
        \end{scope}
        \ifthen{#1=1}{\drawPath{\expPathA}}
        \ifthen{#1=6}{\drawPath[0.5]{\expPathA}}
        \ifthen{#1=2}{\drawPath{\expPathB}}
        \ifthen{#1=6}{\drawPath[0.5]{\expPathB}}
        \ifthen{#1=3}{\drawPath{\expPathC}}
        \ifthen{#1=6}{\drawPath[0.5]{\expPathC}}
        \ifthen{#1=4}{\drawPath{\expPathD}}
        \ifthen{#1=6}{\drawPath[0.5]{\expPathD}}
        \ifthen{#1=5}{\drawPath{\expPathL}}
        \ifthen{#1=5}{\node at (-2.4, 1.8) {\textbf{?}};}
        \ifthen{#1=7}{\drawPath{\expPathE}[detNewPlan]}
        \node[circle,fill=white,draw=black,line width=0.3mm] (s0) at (-3, 0) {{\tiny $s_0$}};
        \node[circle,fill=white,draw=black,line width=0.3mm] (g) at (3, 0) {{\tiny $g$}};
    \end{tikzpicture}
}

\begin{figure*}[ht]
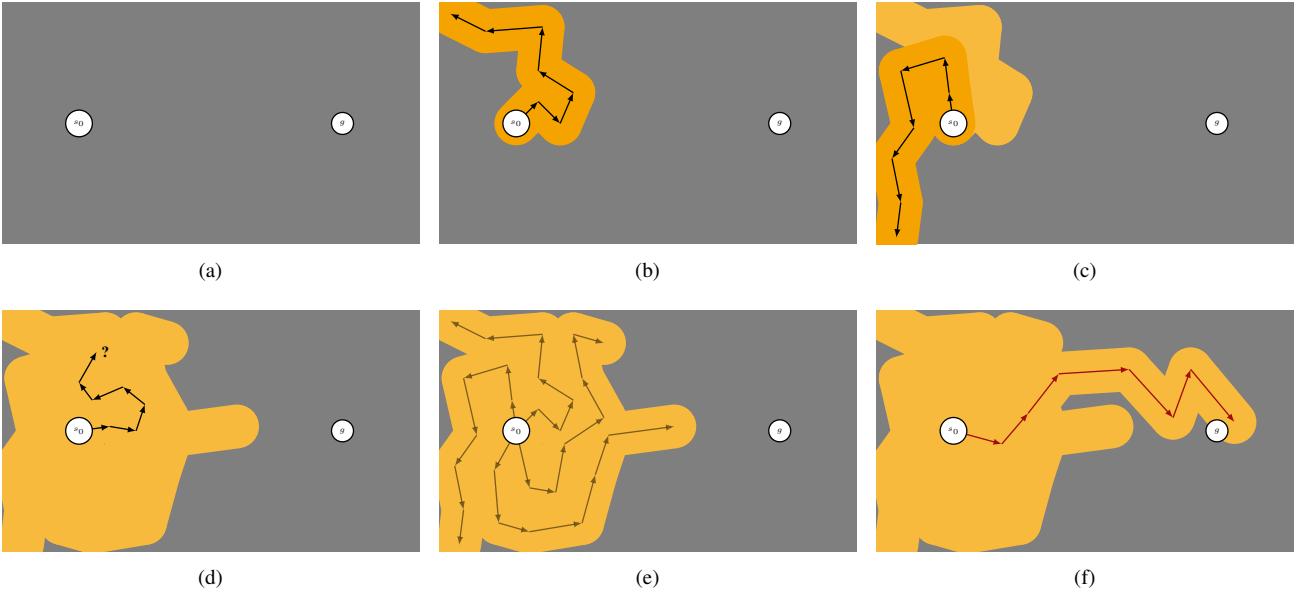

    \centering
    \begin{subfigure}{0.32\linewidth}
        \resizebox{\linewidth}{!}{\drawDet{0}}
        \caption{}
        \label{fig:det_a}
    \end{subfigure}
    \begin{subfigure}{0.32\linewidth}
        \resizebox{\linewidth}{!}{\drawDet{1}}
        \caption{}
        \label{fig:det_b}
    \end{subfigure}
    \begin{subfigure}{0.32\linewidth}
        \resizebox{\linewidth}{!}{\drawDet{2}}
        \caption{}
        \label{fig:det_c}
    \end{subfigure}
    \\[1em]
    \begin{subfigure}{0.32\linewidth}
        \resizebox{\linewidth}{!}{\drawDet{5}}
        \caption{}
        \label{fig:det_d}
    \end{subfigure}
    \begin{subfigure}{0.32\linewidth}
        \resizebox{\linewidth}{!}{\drawDet{6}}
        \caption{}
        \label{fig:det_e}
    \end{subfigure}
    \begin{subfigure}{0.32\linewidth}
        \resizebox{\linewidth}{!}{\drawDet{7}}
        \caption{}
        \label{fig:det_f}
    \end{subfigure}
    \caption{%
        Illustration of the \textit{detachment} problem and our proposed solution to it.
        The agent's objective is to go from the start state $s_0$ to the goal state $g$, but it does not know where $g$ is.
        At the beginning (\subref{fig:det_a}), the entire state space is unexplored, marked by the grey color.
        Driven by the information gain term, the agent starts exploring the state space (\subref{fig:det_b}).
        In the explored areas, marked in yellow, the information gain subsequently decreases, giving the agent less incentive to visit them.
        Hence, the agent avoids these areas in the next episode and continues exploring unknown parts of the state space (\subref{fig:det_c}).
        However, at some point the entire area around the initial state has been explored and the information gain around it becomes close to zero (\subref{fig:det_d}).
        At this point, detachment occurs, since the planner, which relies on local optimization, cannot find a way out of the zero-reward plateau around the initial state.
        We address this issue by maintaining a buffer of past plans (\subref{fig:det_e}) and use those to warm-start the planner.
        Since at least some of the past plans must lead the agent close to the edge of the explored state space, the planner can now find a way out of the explored area and eventually finds the goal (\subref{fig:det_f}).
    }
    \label{fig:detachment}
\end{figure*}

\subsection{Model details and training}
\label{subsec:app_model}

In this section we provide details on the definition and the training of our model.

\subsubsection{Model details}

We compute the means and variances of both the transition model and the reward model with multi-headed dense neural networks.
Specifically, for the transition network, we use two hidden layers with 64 neurons each, followed by a linear head each for the mean and for the variance of the next state.
The reward network has a similar structure, the only difference being that we use only one hidden layer.
As activation function we use leaky ReLU in both networks.
Furthermore, we set the number of ensemble models to 5.

\subsubsection{Training the model}

We follow prior work~\cite{chua2018deep,hafner2019learning} and train our ensemble models individually by maximizing the likelihood of the observed transitions and rewards using a gradient-based optimizer.
A formal downside of this approach is that it neglects the fact that we assume the parameters of our ensemble to be particles of a distribution.
This issue could be fixed by using \gls{svgd}~\cite{liu2016stein}, which treats the parameters of the individual ensemble models as particles of a distribution during training.
Specifically, in \gls{svgd} introduces a \gls{rbf}-kernel term into the loss function to ensure that the particles stay spread out and represent a meaningful distribution.
However, since the number of particles is fairly small compared to their dimension, the probability of two particles collapsing onto one optimum is negligible.
Also, empirically we found that the kernel term has no notable influence on the training.
Hence, we decided not to use \gls{svgd} during model training, as it has no influence on the spread of the particles in our case.

Thus, the optimization problem for each particle $\theta_i$ is given as
\begin{align}
    \label{eq:model_obj}
    \max_{\theta_i} \quad
    \Ex{\left(\vs_{\tau}, \vr_{\tau}, \va_{\tau}, \vs_{\tau-1}\right) \sim \mathcal{D}}{
        \ln \p{\vs_{\tau}}[\va_\tau, \vs_{\tau - 1}, \theta_i]
        + \ln \p{\vr_\tau}[\vs_{\tau}, \va_\tau, \theta_i]}
\end{align}
where $\mathcal{D}$ is the replay buffer containing transitions observed in the past.

\subsubsection{Multi-step prediction loss}
\label{subsec:app_mspl}

During planning, model roll-outs have to be performed to predict the results of different choices of action sequences $\va_{t+1:T}$.
As stated before, we set the planning horizon to $T = t + 20$, which means that the transition model has to be executed 20 times consecutively to produce a full state sequence $\mathbf{\vs}_{t+1:T}$:
\begin{align}
    \begin{split}
        \vs_{\tau} &\sim
        \mathcal{N} \left( \vs_\tau \mid \mu^{\vs}_\theta \left( \vs_{\tau - 1}, \va_\tau \right), \Sigma^{\vs}_\theta \left( \vs_{\tau - 1}, \va_\tau \right) \right)
        \\
        \vr_{\tau} &\sim
        \mathcal{N} \left( r_\tau \mid \mu^r_\theta \left( \vs_\tau, \va_\tau \right), \sigma^r_\theta \left( \vs_\tau, \va_\tau \right) \right)
    \end{split}
    \quad \forall \tau \in \left\{t + 1, \dots, T \right\}
\end{align}
An issue we faced in our experiments is that these model roll-outs suffer from compounding error already after a couple of steps and the produced outputs in the final states of the sequence often explode.
The reason for this behavior is that the models are trained on real states, but executed on predicted states during the roll-outs.
While this might not seem to be a huge source of error at first glance, it is important to keep in mind that in most environments, there are areas of the state space that the agent cannot visit.
If one of the predicted states falls into such an area due to model error, then the model is evaluated on a state it has never seen and also will never see in the dataset.
Hence, the following prediction will likely also be inaccurate, making it unlikely that the model will recover onto a reasonable trajectory again.
Since these erroneous trajectory predictions can result in large objective values, already a few of them are sufficient to divert the planner away from a good plan.

Hence, we follow prior work~\cite{hafner2019learning} and tackle this issue by augmenting the training data of the transition model with predicted states.
Formally, we define a set of new generative models, called m-step predictive models, which are defined as
\begin{align}
    \label{eq:mspl_model}
    \p[m]{\mathbf{\vs}_{0:\vte}, \mathbf{\vr}_{1:\vte}}[\mathbf{\va}_{1:\vte}, \theta]
    =
    \left(
    \prod_{\tau=m}^{\vte}
    \p{\vs_\tau}[\vs_{\tau - m}, \mathbf{\va}_{\tau - m + 1:\tau}, \theta]
    \right)
    \left(
    \prod_{\tau=1}^{m - 1}
    \pr{\mathcal{U}}{\vs_\tau}[l_\vs, u_\vs]
    \right)
    \left(
    \prod_{\tau=1}^{\vte}
    \p{\vr_\tau}[\vs_{\tau}, \va_{\tau}, \theta]
    \right)
\end{align}
where $\vte$ is the episode length, the m-step transition model is given as
\begin{align}
    \label{eq:mspl_m_step_trans}
    \p{\vs_\tau}[\vs_{\tau - m}, \mathbf{\va}_{\tau - m + 1:\tau}, \theta]
    =
    \int
    \prod_{\tau^\prime = \tau - m + 1}^{\tau} \p{\vs_{\tau^\prime}}[\vs_{\tau^\prime - 1}, \va_{\tau^\prime}, \theta]
    d \mathbf{\vs}_{\tau - m + 1:\tau - 1}
\end{align}
and $\mathcal{U}$ is the uniform distribution with $l_\vs$ and $u_\vs$ being very loose bounds of the state $\vs$ and $\psym$ are the 1-step generative models defined in \cref{eq:model_def}.

Under this distribution, the first $m-1$ states are uniformly distributed, and from the $m$-th state on, the $\tau - m$-th state is used to predict the $\tau$-th state.
Intuitively, when predicting the next state $\vs_\tau$, the agent does not use information from the previous $m-1$ states, but rather entirely relies on the action trajectory $\mathbf{\va}_{\tau - m:\tau}$ and the state $\vs_{\tau - m}$ it observed $m$ steps ago.
For this reason, the first $m - 1$ states have to be uniformly distributed, as this model has no access to states before $\vs_0$, which it would require to predict those.
We choose the bounds of the uniform distribution to be very loose, so we can ignore their terms in the loss function completely.
The advantage of such a generative model is that when we use it to train the transition model $\p{\vs_{\tau^\prime}}[\vs_{\tau^\prime - 1}, \va_{\tau^\prime}, \theta]$, the agent learns to make accurate predictions $m$ steps into the future while relying solely on its internal model to generate the intermediate states.

Replacing the 1-step generative model in the objective function in \cref{eq:model_obj} with $\psym_m$ yields the m-step dynamics model learning objective
\begin{equation}
    f_m^{\vs} \left( \mathbf{\vs}_{0:\vte}, \mathbf{\vr}_{1:\vte}, \mathbf{\va}_{1:\vte}, \theta \right)
    \coloneqq
    \sum_{\tau = m}^{\vte}
    \Ex{
        \vs_\tau \sim \p{\hat{\vs}_{\tau - 1}}[\vs_{\tau - m}, \mathbf{\va}_{\tau - m + 1:\tau}, \theta]
    }{
        \ln \p{\vs_{\tau}}[\va_\tau, \vs_{\tau - 1}, \theta]
    }
\end{equation}
To obtain a stochastic gradient of this objective, we compute a \gls{mc} approximation and obtain
\begin{align}
    \begin{split}
        f_m^{\vs} \left( \mathbf{\vs}_{0:\vte}, \mathbf{\vr}_{1:\vte}, \mathbf{\va}_{1:\vte}, \theta \right)
        \approx
        \sum_{\tau = m}^{\vte}
        \ln \p{\vs_{\tau}}[\va_\tau, \hat{\vs}^m_{\tau - 1}, \theta]
    \end{split}
\end{align}
where $\hat{\vs}^m_{\tau - 1}$ is a result of sampling $m$ steps from the transition model:
\begin{align}
    \label{eq:mspl_sampling}
    \begin{split}
        \hat{\vs}^1_{\tau} &\coloneqq \vs_{\tau}
        \\
        \hat{\vs}^i_{\tau} &\sim \mathcal{N}\!\left(
        \mu^{\vs}_\theta \left(\hat{\vs}^{i - 1}_{\tau - 1}, \va_\tau \right),
        \Sigma^{\vs}_\theta \left(\hat{\vs}^{i - 1}_{\tau - 1}, \va_\tau \right)
        \right).
    \end{split}
\end{align}

Like \cite{hafner2019learning}, we stop gradients from flowing through the transition model more than once per sample, as we want model evaluations at later time steps to correct errors made at earlier steps and not vice versa.

To ensure that the model is trained for all step distances it encounters during planning, we define the final dynamics model objective function as a weighted mean over all step distances within the planning horizon:
\begin{align}
    \label{eq:dyn_msp}
    F_H^{\vs} \left( \mathbf{\vs}_{0:\vte}, \mathbf{\vr}_{1:\vte}, \mathbf{\va}_{1:\vte}, \theta \right)
    =
    \sum_{m = 1}^H
    \beta_m
    f_m^{\vs} \left( \mathbf{\vs}_{0:\vte}, \mathbf{\vr}_{1:\vte}, \mathbf{\va}_{1:\vte}, \theta \right)
\end{align}
where $H$ is the relative planning horizon and $\beta$ allows to weight the different step lengths of the objective function.
In our implementation, we choose $\beta_1 \coloneqq 0.5$ and $\beta_i \coloneqq \frac{1}{2(H - 1)}$ for all $i \neq 1$.

Finally, replacing the dynamics part of the maximum likelihood objective (\cref{eq:model_obj}) by the multi-step objective (\cref{eq:dyn_msp}) yields the objective
\begin{align}
    \label{eq:model_obj_msp}
    \max_{\theta_i} \quad
    \Ex{\left(\mathbf{\vs}_{0:\vte}, \mathbf{\vr}_{1:\vte}, \mathbf{\va}_{1:\vte}, \vs_{\tau-1}\right) \sim \mathcal{D}}{
        F_H^{\vs} \left( \mathbf{\vs}_{0:\vte}, \mathbf{\vr}_{1:\vte}, \mathbf{\va}_{1:\vte}, \theta_i \right)
        + \sum_{\tau = 1}^\vte
        \ln \p{\vr_\tau}[\vs_{\tau}, \va_\tau, \theta_i]}.
\end{align}

\subsection{Hardening the reward model via multi-step prediction loss}
\label{subsec:mspl_rew}

One issue of the multi-step prediction loss as proposed by \cite{hafner2019learning} is that it does not train the reward model on predicted data.
While, unlike the dynamics model, the accuracy of the reward model does not suffer from compounding model error, it will still get evaluated on states predicted by the dynamics model, which might be inaccurate.
Hence, during planning, the reward model will be evaluated on states it was not trained on and, thus, might make predictions that are arbitrarily wrong.
Although we found the effect of this issue on the planner to be less severe than the effect of compounding state transition error, it still has a notable impact on the performance of the planner.
Thus, in this section, we take the idea of \cite{hafner2019learning} one step further and extend the multi-step prediction objective to the reward model.

We start by rewriting \cref{eq:mspl_model} to condition the reward model on predicted states instead of observed states:
\begin{align}
    \begin{split}
        \p[m]{\mathbf{\vs}_{0:\vte}, \mathbf{\vr}_{1:\vte}}[\mathbf{\va}_{1:\vte}, \theta]
        =&
        \left(
        \prod_{\tau=m}^{\vte}
        \p{x_\tau}[\vs_{\tau - m}, \mathbf{\va}_{\tau - m + 1:\tau}, \theta]
        \right)
        \left(
        \prod_{\tau=1}^{m - 1}
        \mathcal{U} \left( \vs_\tau, l_{\vs}, u_{\vs} \right)
        \right)
        \\
        &\left(
        \prod_{\tau=m}^{\vte}
        \p{\vr_\tau}[\vs_{\tau - m + 1}, \mathbf{\va}_{\tau - m + 1:\tau}, \theta]
        \right)
        \left(
        \prod_{\tau=1}^{m - 1}
        \mathcal{U} \left( \vr_\tau, l_\vr, u_\vr \right)
        \right)
    \end{split}
\end{align}
where $l_\vr$ and $u_\vr$ are again set to be very loose bounds, such that we can neglect the impact of the uniform distribution on the loss, and
\begin{align}
    \p{\vr_\tau}[\vs_{\tau - m + 1}, \mathbf{\va}_{\tau - m + 1:\tau}, \theta]
    =
    \int
    \p{\vr_\tau}[\vs_{\tau}, \va_{\tau}, \theta]
    \prod_{ \mathclap{ \tau^\prime = \tau - m + 2 }}^{\tau}
    \p{\vs_{\tau^\prime}}[\vs_{\tau^\prime - 1}, \va_{\tau^\prime}, \theta]
    d \mathbf{\vs}_{\tau - m + 2:\tau - 1}.
\end{align}
Similar to the $m$-step dynamics model, this model only considers the state $m$ steps ago and the action trajectory for computing the reward.

Analogously to the $m$-step dynamics model, the \gls{mc} approximation of the $m$-step prediction objective function for the reward is given as
\begin{align}
    \begin{split}
        f_m^{\vr} \left( \mathbf{\vs}_{0:\vte}, \mathbf{\vr}_{1:\vte}, \mathbf{\va}_{1:\vte}, \theta \right)
        \approx
        \sum_{\tau = m}^{\vte}
        \ln \p{\vr_\tau}[\hat{\vs}^m_{\tau}, \va_{\tau}, \theta]
    \end{split}
\end{align}
where $\hat{\vs}^m_{\tau}$ is again sampled according to \cref{eq:mspl_sampling}.

Thus, the overall objective function is given as
\begin{align}
    F_H
    \coloneqq
    \Ex{\left(\mathbf{\vs}_{0:\vte}, \mathbf{\vr}_{1:\vte}, \mathbf{\va}_{1:\vte}, \vs_{\tau-1}\right) \sim \mathcal{D}}{
        F_H^{\vs} \left( \mathbf{\vs}_{0:\vte}, \mathbf{\vr}_{1:\vte}, \mathbf{\va}_{1:\vte}, \theta_i \right)
        + F_H^{\vr} \left( \mathbf{\vs}_{0:\vte}, \mathbf{\vr}_{1:\vte}, \mathbf{\va}_{1:\vte}, \theta_i \right)
    }
\end{align}
where
\begin{align}
    F_H^{\vr} \left( \mathbf{\vs}_{0:\vte}, \mathbf{\vr}_{1:\vte}, \mathbf{\va}_{1:\vte}, \theta \right)
    =
    \sum_{m = 1}^H
    \beta_m
    f_m^{\vr} \left( \mathbf{\vs}_{0:\vte}, \mathbf{\vr}_{1:\vte}, \mathbf{\va}_{1:\vte}, \theta \right).
\end{align}

After simplifying, we obtain the optimization problem
\begin{align}
    \label{eq:model_obj_rmsp}
    \max_{\theta_i} \quad
    \Ex{\left(\mathbf{\vs}_{0:\vte}, \mathbf{\vr}_{1:\vte}, \mathbf{\va}_{1:\vte}, \vs_{\tau-1}\right) \sim \mathcal{D}}{
        \sum_{m = 1}^H
        \beta_m
        \sum_{\tau = m}^{\vte}
        \ln \p{\vr_\tau}[\hat{\vs}^m_{\tau - 1}, \va_\tau, \theta_i]
        +
        \ln \p{\vr_\tau}[\hat{\vs}^m_{\tau}, \va_{\tau}, \theta_i]
    }
\end{align}
where $\hat{\vs}_{\tau}$ is defined as in \cref{eq:mspl_sampling}.

We use this optimization problem for model learning throughout all our experiments.

\subsubsection{Optimization details}
We use Adam~\cite{kingma2014adam} as optimizer, which we run after every episode for 20 steps per transition in the replay buffer, but not for more than 600 steps.
Furthermore, we choose the replay buffer size such that all transitions that the agent will encounter during the training will fit.
That means that no sample will ever be deleted and catastrophic forgetting can effectively not occur.

\subsection{The exploration-exploitation trade-off: adjusting $\beta$}

A crucial hyperparameter that must be tuned is the weight of the intrinsic term $\beta$, as it is responsible for controlling the exploration-exploitation trade-off.
In theory the intrinsic term should go to zero over time and, thus, the agent should be driven more by the extrinsic term in the later stages of the training.
However, in practice, we found the intrinsic term can stay non-zero for a very long time, due to approximation errors and the general difficulty of learning an accurate model of a manipulation task.
Hence, $\beta$ needs to be adjusted carefully to ensure that the agent explores properly in the beginning and focuses on exploitation in the later states of the training, when the goal has been understood.

For the tasks presented in \cref{sec:exp}, we found that by setting $\beta$ to the constant $10^6$ yields reasonable results if the intrinsic term is \glsxtrlong{mi}.
However, it is likely that $\beta$ has to be tuned differently for other tasks.

Adjusting $\beta$ for \glsxtrlong{li}, on the other hand, turned out to be more challenging.
While a constant value of $2 \cdot 10^5$ is sufficient to solve the \textit{Tilted Pushing} task, the performance on the \textit{Tilted Pushing Maze} task with the same value is poor.
By a hyperparameter search we found that there seems to be no constant $\beta$ that ensures that the agent is exploring properly in the beginning, while exploiting properly in the end for this task.
We alleviate this issue by choosing $\beta$ adaptively, depending on the reward that has been achieved in the roll-outs so far.
The intuition behind this idea is that as the agent obtains higher reward, it should focus more on exploitation, as it has obviously found a way to solve the task.
Specifically, we set
\begin{align}
    \beta \coloneqq \alpha f \left( \vr_1, \dots, \vr_N \right) + \gamma
\end{align}
where $\vr_1, \dots, \vr_N$ are the per-step rewards the agent obtained during the course of the training (not only the current roll-out, but the entire training), $f$ is an aggregation function for the reward and $\alpha, \gamma$ are hyperparameters for the scaling and offset.
We looked into two choices for $f$: the maximum and the running average.
Empirically, we found that the best results are obtained on the \textit{Tilted Pushing Maze} task by using $f \coloneqq \max$, $\alpha \coloneqq 10^8$, and $\gamma \coloneqq 2 \cdot 10^5$.

However, how to set $\beta$ optimally for a variety of tasks remains an open research question that we plan to address in future work.

\subsection{Task details}

In all our experiments, the agent is restricted to move the finger in a 2D plane above the table and rotate it up to 0.3 rad.
The linear movement of the finger is controlled by specifying a velocity difference, that a low-level controller will then try to realize.
Utilizing a velocity difference here instead of a target velocity directly yields the advantage that jerky movements can be avoided by simply restricting the action space.
To ensure that the robot cannot leave the table to the side, the low level controller will remove any components of the target velcity vector that lead the agent outside the table bounds.
The finger rotation is controlled by specifying a target angular velocity, which is again realized by a low-level controller.
This low level controller also makes sure that the angular limits are not breached.
As noted before, the control frequency of the high-level controller is 4 Hz.

The agent fully observes a 10 dimensional state vector, which comprises the 2D finger position and velocity, the 1D finger rotation and velocity, and the 2D ball position and velocity.
Note that the spin of the ball is not observed, since we are not able to observe this property on the real robot.
However, we found that observing the spin of the ball is not necessary for solving this task.

The per-step reward provided by the environment is fully deterministic and given as
\begin{align}
    \vr \left( \vs_t, \va_t \right) = 0.001 \lVert \va_t \rVert^2 + g \left( \vs_t \right)
\end{align}
where $\va_t$ is the action vector normalized to lie within $[-1, 1]$ and $g \colon \R^{N_\vs} \rightarrow \{0, 1\}$ is a function that is $1$ when the projection of the ball center on the table is within the target area and $0$ otherwise.

Both in simulation and in the real world experiments, we use a ball with radius 2cm and a gripper width of 2cm.
The target zone begins 8cm below the top of the table and is 8cm $\times$ 5cm large.
Our robot is a UR10.

In the following we briefly highlight some details about the simulated and real world experiments.

\subsubsection{Simulation}

In simulation we use a table size of 50cm $\times$ 57cm and a table inclination of 0.2 rad.
To help with the maneuvering of the ball around the holes, we set the friction between the ball and the finger slightly higher in the \textit{Tilted Pushing Maze} task, compared to the \textit{Tilted Pushing} task.

\subsubsection{Real world}

In the real world, we use a table size of approximately 54cm x 518cm.
As visible in \cref{fig:tasks_real}, the table and the barriers are set up in such a way that the ball always rolls back to the same position at the bottom of the table if it is dropped.
From this position it can be picked up autonomously by the robot and pushed up to the starting position.
To ensure that the ball is not dropped on the way to the starting position, we cut a rail into the table that stabilizes the ball until the starting position is reached.
In case the ball gets stuck somewhere on the bottom barrier and does not arrive in the center, we implemented an autonomous ball recovery procedure, in which the robot moves the finger along the barrier to knock the ball back into the center.
This setup allows us to perform our experiments completely autonomously and unattended, regularly reaching up to 25 hours of uninterrupted training without intervention.

To make the balancing of the ball at our control frequency of 4 Hz possible, we add some rubber tape to the fingertip, which gives it a bit of extra friction.
Furthermore, we track the ball using an Optitrack system with 9 Flex 13 cameras.
Since Optitrack only estimates the position of objects and not their velocity, we implement a Kalman filter to estimate it.

\printbibliography[keyword=appendix,title=Appendix References]